\documentclass[10pt,twocolumn,letterpaper]{article}

\usepackage{amsmath,amsfonts,bm}

\def\eqref#1{equation~\ref{#1}}

\def\1{\bm{1}}

\def\mJ{{\bm{J}}}

\def\mL{{\bm{L}}}

\DeclareMathAlphabet{\mathsfit}{\encodingdefault}{\sfdefault}{m}{sl}
\SetMathAlphabet{\mathsfit}{bold}{\encodingdefault}{\sfdefault}{bx}{n}

\def\gD{{\mathcal{D}}}

\usepackage{cvpr}

\usepackage{graphicx}
\usepackage{amsmath}
\usepackage{amssymb}
\usepackage{booktabs}
\usepackage{multirow}
\usepackage{pifont}
\usepackage{nicefrac}
\usepackage{bbding}
\usepackage[pagebackref,breaklinks,colorlinks]{hyperref}

\mathchardef\mhyphen="2D

\usepackage[capitalize]{cleveref}
\crefname{section}{Sec.}{Secs.}
\Crefname{section}{Section}{Sections}
\Crefname{table}{Table}{Tables}
\crefname{table}{Tab.}{Tabs.}

\newcommand{\red}[1]{{\color{red}{#1}}}

\newcommand{\tablestyle}[2]{\setlength{\tabcolsep}{#1}\renewcommand{\arraystretch}{#2}\centering\footnotesize}
\newlength\savewidth\newcommand\shline{\noalign{\global\savewidth\arrayrulewidth
		\global\arrayrulewidth .8pt}\hline\noalign{\global\arrayrulewidth\savewidth}}

\def\cvprPaperID{2061} 
\def\confName{CVPR}
\def\confYear{2022}

\usepackage{tabulary}
\usepackage{overpic}
\usepackage{xcolor}
\usepackage{colortbl}
\usepackage{caption}
\usepackage{subcaption}
\usepackage{wrapfig}
\usepackage{comment}
\usepackage{url}

\def\x{$\times$}
\def\sqr{$^2$}
\def\eg{\textit{e.g.}}
\def\ie{\textit{i.e.}}

\definecolor{b0}{RGB}{175,205,238}
\definecolor{b1}{RGB}{145,177,221}
\definecolor{b2}{RGB}{114,148,203}
\definecolor{b3}{RGB}{85,121,187}

\newcolumntype{x}[1]{>{\centering\arraybackslash}p{#1pt}}
\newcommand{\blockctd}[2]{\([\text{3$\times$3$^\text{2}$, #1}]\)$\times$#2}
\newcommand{\blockxtd}[3]{\(\left[\begin{array}{c}\text{1$\times$1$^\text{2}$, #2}\\[-.1em] \text{3$\times$3$^\text{2}$, #2}\\[-.1em] \text{1$\times$1$^\text{2}$, #1}\end{array}\right]\)$\times$#3}
\newcommand{\myblocks}[3]{\(\left[\begin{array}{c}\text{1$\times$1$^\text{2}$, #2}\\[-.1em] \text{1$\times$3$^\text{2}$, #2}\\[-.1em] \text{1$\times$1$^\text{2}$, #1}\end{array}\right]\)$\times$#3}
\newcommand{\myblockt}[3]{\(\left[\begin{array}{c}\text{\underline{3$\times$1$^\text{2}$}, #2}\\[-.1em] \text{1$\times$3$^\text{2}$, #2}\\[-.1em] \text{1$\times$1$^\text{2}$, #1}\end{array}\right]\)$\times$#3}
\newcommand{\blocks}[3]{\multirow{3}{*}{\(\left[\begin{array}{c}\text{1$\times$1$^\text{2}$, #2}\\[-.1em] \text{1$\times$3$^\text{2}$, #2}\\[-.1em] \text{1$\times$1$^\text{2}$, #1}\end{array}\right]\)$\times$#3}
}
\newcommand{\blockt}[3]{\multirow{3}{*}{\(\left[\begin{array}{c}\text{\underline{3$\times$1$^\text{2}$}, #2}\\[-.1em] \text{1$\times$3$^\text{2}$, #2}\\[-.1em] \text{1$\times$1$^\text{2}$, #1}\end{array}\right]\)$\times$#3}
}

\title{Revisiting Skeleton-based Action Recognition}

\author{Haodong Duan$^{1,3}$ \hspace{5mm} Yue Zhao$^{2}$ \hspace{5mm} Kai Chen$^{3, 5}$ \hspace{5mm} Dahua Lin$^{1,3}$ \hspace{5mm} Bo Dai$^{3,4}$ \ \Envelope \\
$^{1}$The Chinese University of HongKong \hspace{8mm}  $^{2}$The University of Texas at Austin \\ 
$^{3}$Shanghai AI Laboratory \hspace{4mm} $^{4}$S-Lab, Nanyang Technological University \hspace{4mm} $^{5}$SenseTime Research }

\begin{document}
\maketitle

\begin{abstract}
\vspace{-2mm}

Human skeleton, as a compact representation of human action, has received increasing attention in recent years.
Many skeleton-based action recognition methods adopt GCNs to extract features on top of human skeletons.
Despite the positive results shown in these attempts,
GCN-based methods are subject to limitations in robustness, interoperability, and scalability.
In this work, we propose PoseConv3D, a new approach to skeleton-based action recognition.
PoseConv3D relies on a 3D heatmap volume instead of a graph sequence as the base representation of human skeletons.
Compared to GCN-based methods,
PoseConv3D is more effective in learning spatiotemporal features,
more robust against pose estimation noises,
and generalizes better in cross-dataset settings.
Also, PoseConv3D can handle multiple-person scenarios without additional computation costs.
The hierarchical features can be easily integrated with other modalities at early fusion stages,
providing a great design space to boost the performance.
PoseConv3D achieves the state-of-the-art on five of six standard skeleton-based action recognition benchmarks.
Once fused with other modalities, it achieves the state-of-the-art on all eight multi-modality action recognition benchmarks. 
Code has been made available at: \url{https://github.com/kennymckormick/pyskl}. 
\end{abstract}

\vspace{-8mm}

\section{Introduction}

Action recognition is a central task in video understanding.
Existing studies have explored various modalities for feature representation,
such as RGB frames \cite{wang2016temporal,tran2015learning,carreira2017quo},
optical flows \cite{simonyan2014two},
audio waves \cite{xiao2020audiovisual},
and human skeletons \cite{yan2018spatial,weinzaepfel2021mimetics}.
Among these modalities,
skeleton-based action recognition
has received increasing attention in recent years due to its action-focusing nature and compactness.
In practice, human skeletons in a video are mainly represented as
a sequence of joint coordinate lists, where the coordinates are extracted by pose estimators.
Since only the pose information is included,
skeleton sequences capture only action information
while being immune to contextual nuisances, such as background variation and lighting changes.

\begin{figure}[t]
	\captionsetup{font=small}
	\captionsetup[subfloat]{font=footnotesize}
	\centering
	\resizebox{\linewidth}{!}{
	\subfloat[2D poses estimated with HRNet. \label{fig-suba}]{
		\hspace{-2mm}
		\includegraphics[width=.45\linewidth]{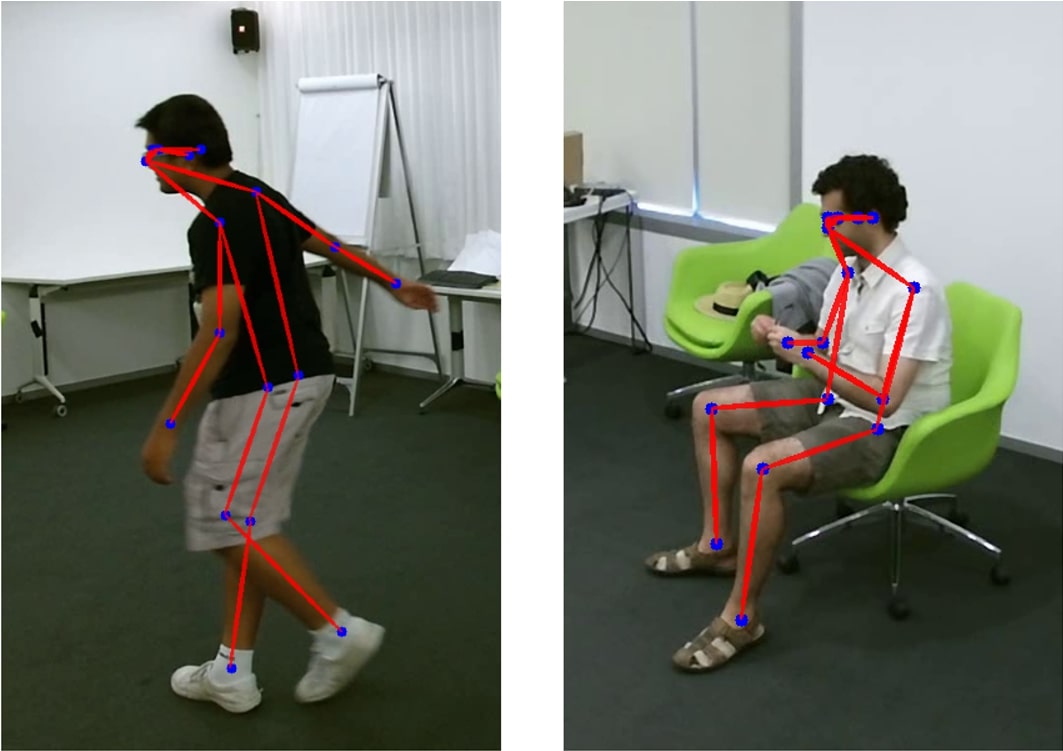}
		\hspace{1mm}
		\includegraphics[width=.45\linewidth]{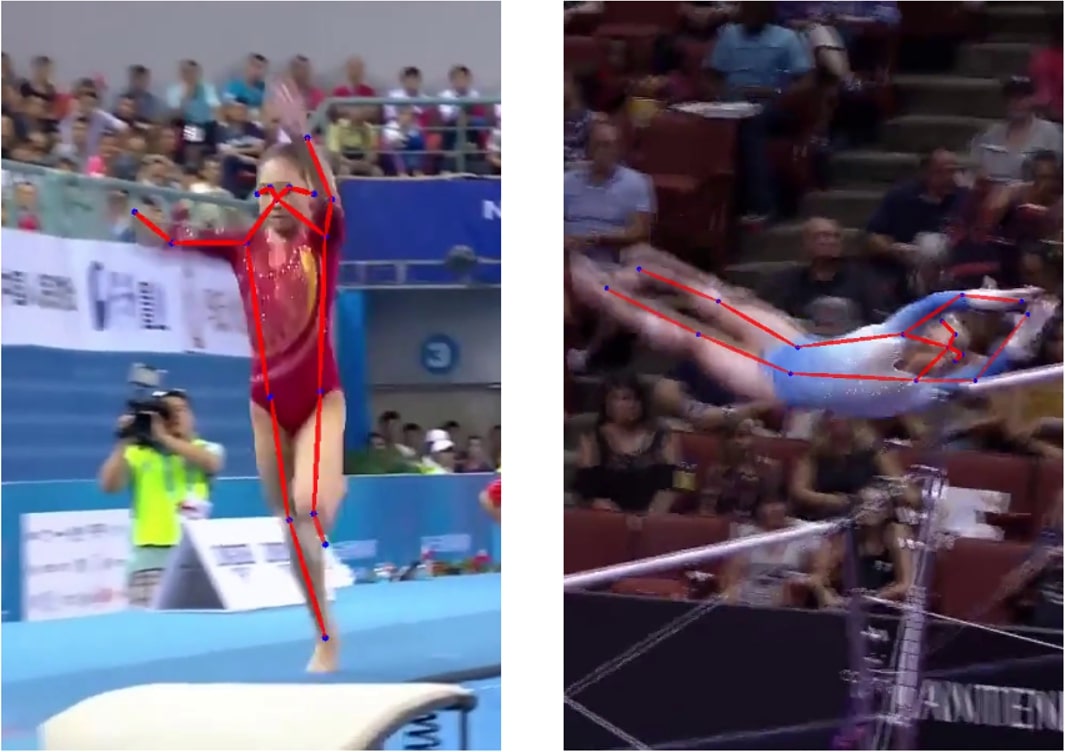}
		\vspace{-.5mm}
	}}
	\vspace{1.5mm}
	\centering
	\resizebox{\linewidth}{!}{
	\hspace{-2mm}
	\subfloat[3D poses collected with Kinect. \label{fig-subb}]{
		\includegraphics[width=.45\linewidth]{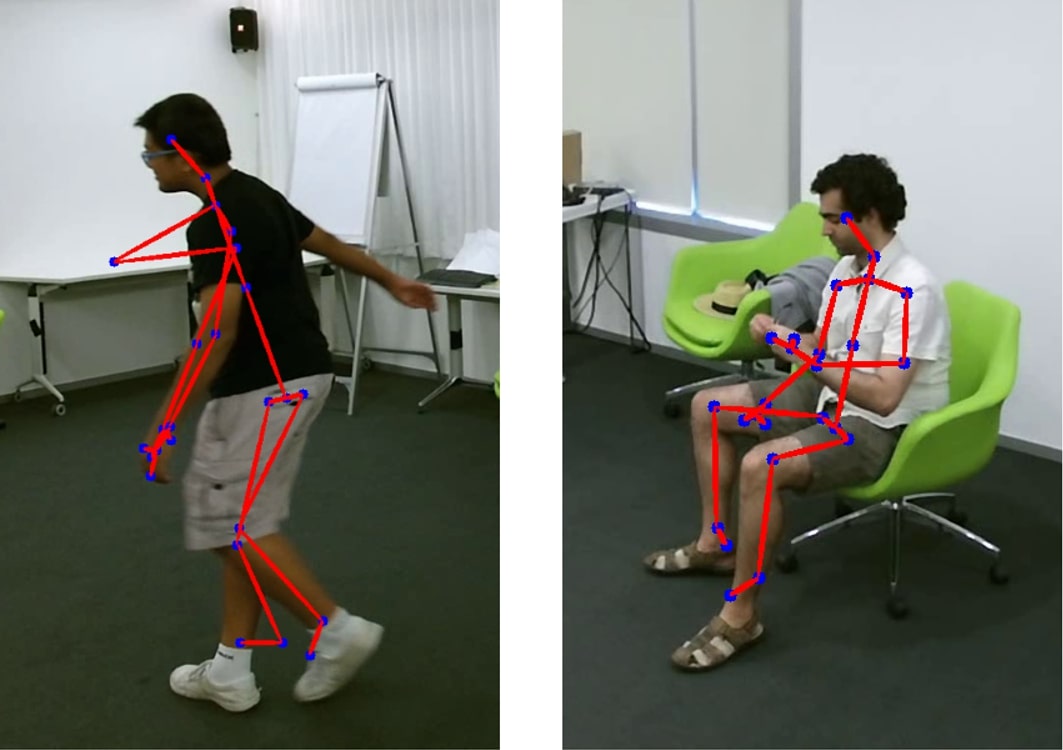}
		\vspace{-.5mm}
	}
	\hspace{1mm}
	\subfloat[3D poses estimated with VIBE. \label{fig-subc}]{
		\includegraphics[width=.45\linewidth]{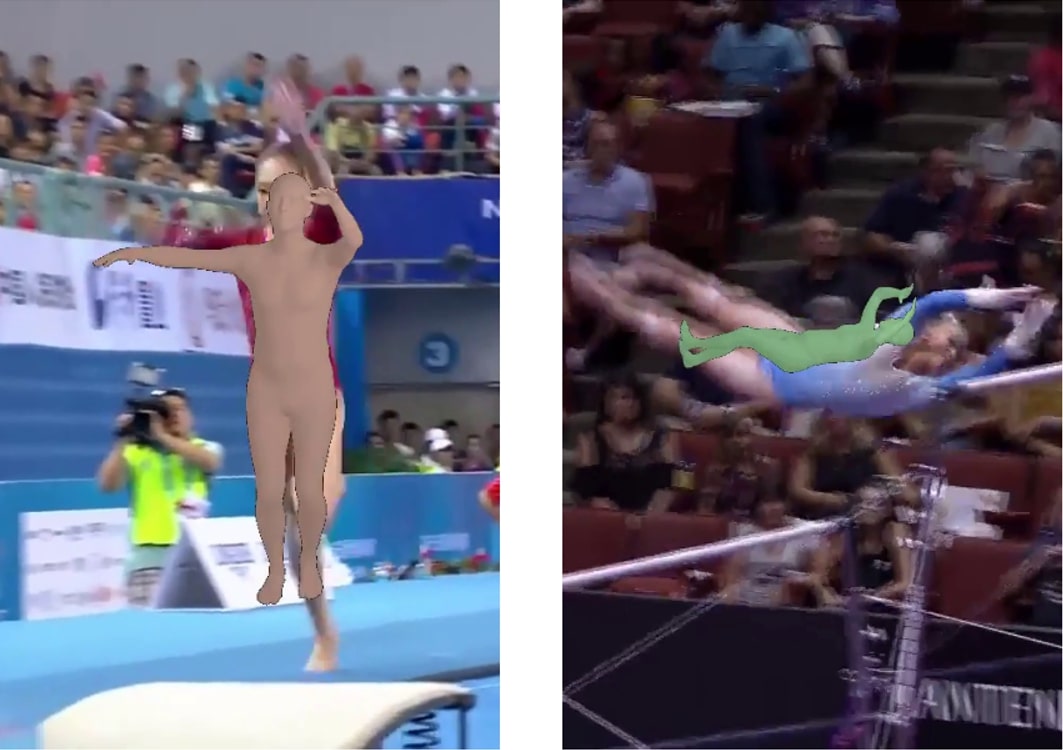}
		\vspace{-.5mm}
	}
	}
	\vspace{-7mm}
	\caption{\textbf{PoseConv3D takes 2D poses as inputs. } 
		In general, 2D poses are of better quality than 3D poses. 
		We visualize 2D poses estimated with HRNet for videos in NTU-60 and FineGYM in (a). 
		Apparently, their quality is much better than 3D poses collected by sensors (b) or estimated with state-of-the-art estimators (c). }
	\label{fig-2Dvs3D}
	\vspace{-2mm}
\end{figure}

\begin{table}[t]
	\captionsetup{font=small, position=top}
	\captionsetup[subfloat]{font=footnotesize, position=top}
	\centering
	\caption{\textbf{Differences between PoseConv3D and GCN.}}
	\label{tab-diff}
	\vspace{-2mm}
	\resizebox{.9\linewidth}{!}{
		\tablestyle{4pt}{1.1}
		\footnotesize
		\begin{tabular}{ccc} 
			\shline
			& Previous Work  & PoseConv3D        \\ 
			\shline
			Input        & 2D / 3D Skeleton   & 2D Skeleton \\ 
			Format       & Coordinates          & 3D Heatmap Volumes  \\ 
			Architecture & GCN                  & 3D-CNN        \\  
			\shline
		\end{tabular}
	}
	\vspace{-4mm}
\end{table}

Among all the methods for skeleton-based action recognition \cite{du2015hierarchical,wang2012mining,vemulapalli2014human},
graph convolutional networks (GCN)~\cite{yan2018spatial} have been one of the most popular approaches.
Specifically,
GCNs regard every human joint at every timestep as a node.
Neighboring nodes along the spatial and temporal dimensions are connected with edges.
Graph convolution layers are then applied to the constructed graph
to discover action patterns across space and time.
Due to the good performance on standard benchmarks for skeleton-based action recognition,
GCNs have been a standard approach when processing skeleton sequences.

While encouraging results have been observed,
GCN-based methods are limited in the following aspects:
(1) \emph{Robustness:}
While GCN directly handles coordinates of human joints,
its recognition ability is significantly affected
by the distribution shift of coordinates,
which can often occur when applying a different pose estimator to acquire the coordinates.
A small perturbation in coordinates often leads to completely different predictions~\cite{zhu2019robust}.
(2) \emph{Interoperability:}
Previous works have shown that representations from different modalities,
such as RGB, optical flows, and skeletons, are complementary.
Hence, an effective combination of such modalities can often
result in a performance boost in action recognition.
However, GCN is operated on an irregular graph of skeletons, making it difficult
to fuse with other modalities that are often represented on regular grids,
especially in the early stages.
(3) \emph{Scalability:}
In addition,
since GCN regards every human joint as a node,
the complexity of GCN scales linearly with the number of persons,
limiting its applicability to scenarios that involve multiple persons,
such as group activity recognition.

In this paper, we propose a novel framework \textbf{PoseConv3D}
that serves as a competitive alternative to GCN-based approaches.
In particular,
PoseConv3D takes as input 2D poses obtained by modern pose estimators shown in Figure~\ref{fig-2Dvs3D}.
The 2D poses are represented by stacks of heatmaps of skeleton joints rather than coordinates operated on a human skeleton graph.
The heatmaps at different timesteps will be stacked along the temporal dimension to form a 3D heatmap volume.
PoseConv3D then adopts a 3D convolutional neural network on top of the 3D heatmap volume
to recognize actions.
Main differences between PoseConv3D and GCN-based approaches are summarized in Table~\ref{tab-diff}.

PoseConv3D can address the limitations of GCN-based approaches stated above.
First, using 3D heatmap volumes is more robust to the up-stream pose estimation:
we empirically find that PoseConv3D generalizes well across input skeletons obtained by different approaches.
Also, PoseConv3D, which relies on heatmaps of the base representation,
enjoys the recent advances in convolutional network architectures
and is easier to integrate with other modalities
into multi-stream convolutional networks.
This characteristic opens up great design space to further improve the recognition performance.
Finally, PoseConv3D can handle different numbers of persons without increasing computational overhead
since the complexity over 3D heatmap volume is independent of the number of persons.
To verify the efficiency and effectiveness of PoseConv3D,
we conduct comprehensive studies across several datasets,
including FineGYM~\cite{shao2020finegym}, NTURGB-D~\cite{Liu_2019_NTURGBD120}, UCF101~\cite{soomro2012ucf101},
HMDB51~\cite{kuehne2011hmdb}, Kinetics400~\cite{carreira2017quo}, and Volleyball~\cite{ibrahim2016hierarchical},
where PoseConv3D achieves state-of-the-art performance compared to GCN-based approaches.

\section{Related Work}

\noindent
\textbf{3D-CNN for RGB-based action recognition. }
3D-CNN is a natural extension of 2D-CNN for spatial feature learning to spatiotemporal in videos.
It has long been used in action recognition \cite{ji20123d, tran2015learning}. 
Due to a large number of parameters, 
3D-CNN requires huge amounts of videos to learn good representation.
3D-CNN has become the mainstream approach for action recognition 
since I3D~\cite{carreira2017quo}.
From then on, many advanced 3D-CNN architectures~\cite{tran2018closer, feichtenhofer2019slowfast, tran2019video, feichtenhofer2020x3d} have been proposed by the action recognition community,
which outperform I3D both in precision and efficiency. 
In this work, we first propose to use 3D-CNN with 3D heatmap volumes as inputs
and obtain the state-of-the-art in skeleton-based action recognition.

\noindent
\textbf{GCN for skeleton-based action recognition. }
Graph convolutional network is widely adopted in skeleton-based action recognition~\cite{yan2018spatial, song2020stronger, gupta2021quo, chen2021channel, song2021constructing, cai2021jolo}.
It models human skeleton sequences as spatiotemporal graphs.
ST-GCN~\cite{yan2018spatial} is a well-known baseline for GCN-based approaches, 
which combines spatial graph convolutions and interleaving temporal convolutions 
for spatiotemporal modeling.
Upon the baseline, adjacency powering is used for multiscale modeling~\cite{liu2020disentangling, li2019actional},
while self-attention mechanisms improve the modeling capacity~\cite{shi2019two, li2019spatio}. 
Despite the great success of GCN in skeleton-based action recognition, it is also limited in robustness~\cite{zhu2019robust} and scalability.
Besides, for GCN-based approaches, 
fusing features from skeletons and other modalities may need careful design~\cite{das2020vpn}. 

\noindent
\textbf{CNN for skeleton-based action recognition. }
Another stream of work adopts convolutional neural networks for skeleton-based action recognition. 
2D-CNN-based approaches first model the skeleton sequence as a pseudo image 
based on manually designed transformations.
One line of works aggregates heatmaps along the temporal dimension into a 2D input 
with color encodings~\cite{choutas2018potion} or learned modules~\cite{yan2019pa3d, asghari2020dynamic}.
Although carefully designed, information loss still occurs during the aggregation, which leads to inferior recognition performance.
Other works~\cite{ke2017new,luvizon20182d,caetano2019skelemotion, li2018co, joze2020mmtm} directly convert the coordinates in a skeleton sequence to a pseudo image with transformations,
typically generate a 2D input of shape $K\times T$,
where $K$ is the number of joints, $T$ is the temporal length.
Such input cannot exploit the locality nature of convolution networks,
which makes these methods not as competitive as GCN on popular benchmarks~\cite{caetano2019skelemotion}.
Only a few previous works have adopted 3D-CNNs for skeleton-based action recognition. 
To construct the 3D input, they either stack the pseudo images of distance matrices~\cite{hernandez20173d,lin2020image}
or directly sum up the 3D skeletons into a cuboid~\cite{liu2017two}. 
These approaches also severely suffer from information loss and obtain much inferior performance to the state-of-the-art.
Our work stacks heatmaps along the temporal dimension to form 3D heatmap volumes,
preserving all information during this process.
Besides, we use 3D-CNN instead of 2D-CNN due to its good capability for spatiotemporal feature learning.

\section{Framework}

\begin{figure*}[t]
	\centering
	\captionsetup{font=small}
	\vspace{-2mm}
	\includegraphics[width=.95\linewidth]{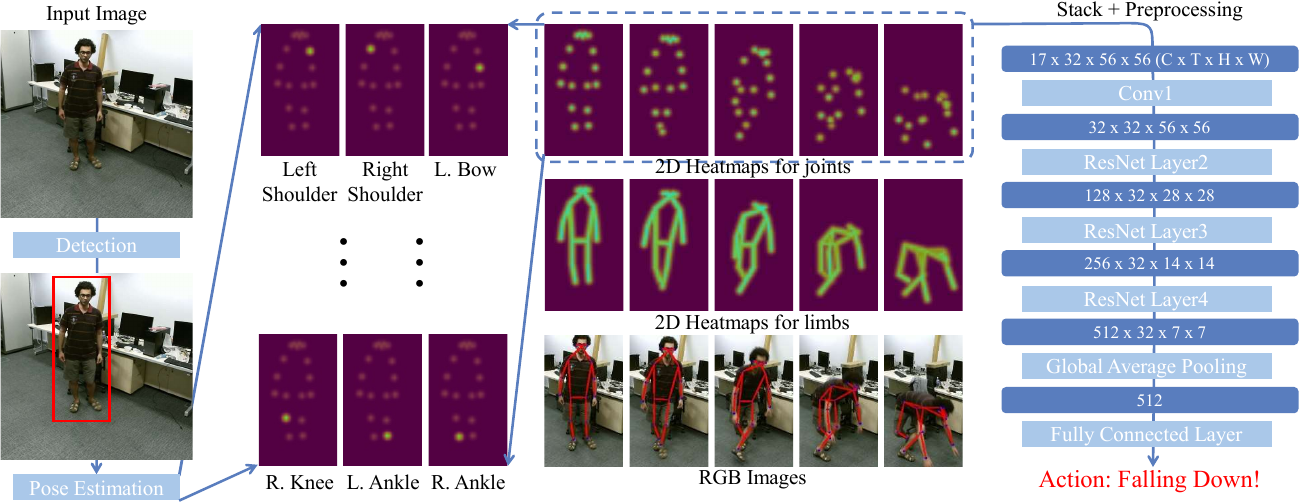}
	\caption{ \textbf{Our Framework. }
		For each frame in a video,
		we first use a two-stage pose estimator (detection + pose estimation) for 2D human pose extraction.
		Then we stack heatmaps of joints or limbs along the temporal dimension and apply pre-processing to the generated 3D heatmap volumes.
		Finally, we use a 3D-CNN to classify the 3D heatmap volumes. }
	\label{fig-framework}
	\vspace{-3mm}
\end{figure*}

We propose \textbf{PoseConv3D},
a \textbf{3D-CNN}-based approach for skeleton-based action recognition,
which can be a competitive alternative to GCN-based approaches,
outperforming GCN under various settings in terms of accuracy with improved robustness, interoperability, and scalability.
An overview of PoseConv3D is depicted in Figure~\ref{fig-framework},
and details of PoseConv3D will be covered in the following sections.
We begin with a review of skeleton extraction,
which is the basis of skeleton-based action recognition but is often overlooked in previous literature.
We point out several aspects that should be considered when choosing a skeleton extractor
and motivate the use of 2D skeletons in PoseConv3D\footnote{
	PoseConv3D can also work with 3D skeletons.
	An example solution is to divide a 3D skeleton $(x,y,z)$ into three 2D skeletons respectively
	using $(x,y), (y,z)$ and $(x,z)$.}.
Subsequently, we introduce 3D Heatmap Volume that is the representation of a 2D skeleton sequence used in PoseConv3D,
followed by the structural designs of PoseConv3D,
including a variant that focuses on the modality of human skeletons
as well as a variant that combines the modalities of human skeletons and RGB frames
to demonstrate the interoperability of PoseConv3D.


\subsection{Good Practices for Pose Extraction}
\label{sec-pose-extraction}
Being a critical pre-processing step for skeleton-based action recognition,
human skeleton or pose extraction largely affects the final recognition accuracy.
However, its importance is often overlooked in previous literature,
in which poses estimated by sensors \cite{Shahroudy_2016_NTURGBD, Liu_2019_NTURGBD120}
or existing pose estimators \cite{8765346, yan2018spatial} are used without considering the potential effects.
Here we conduct a review on key aspects of pose extraction to find a good practice.

In general, 2D poses are of better quality compared to 3D poses, 
as shown in Figure~\ref{fig-2Dvs3D}.
We adopt 2D Top-Down pose estimators~\cite{newell2016stacked,xiao2018simple,sun2019deep} for pose extraction.
Compared to its 2D Bottom-Up counterparts~\cite{newell2017associative,cao2017realtime,cheng2020higherhrnet},
Top-Down methods obtain superior performance on standard benchmarks such as COCO-keypoints~\cite{lin2014microsoft}.
In most cases, we feed proposals predicted by a human detector to the Top-Down pose estimators,
which is sufficient enough to generate 2D poses of good quality for action recognition.
%
%
When only a few persons are of interest out of dozens of candidates \footnote{
In FineGym, there exists dozens of audience, while only the pose of the athlete matters.},
some priors are essential for skeleton-based action recognition to achieve good performance, 
\eg, knowing the interested person locations at the first frame of the video.
In terms of the storage of estimated heatmaps,
they are often stored as coordinate-triplets $(x,y,c)$ in previous literature,
where $c$ marks the maximum score of the heatmap and $(x,y)$ is the corresponding coordinate of $c$.
In experiments, we find that
coordinate-triplets $(x, y, c)$ help save the majority of storage space
at the cost of little performance drop.
The detailed ablation study is included in Sec~\ref{sec-abl-pose-extraction}.

\subsection{From 2D Poses to 3D Heatmap Volumes}
\label{sec-gen_volume}

After 2D poses are extracted from video frames, to feed into PoseConv3D, we reformulate them into a 3D heatmap volume.
Formally, we represent a 2D pose as a heatmap of size $K \times H \times W$,
where $K$ is the number of joints, $H$ and $W$ are the height and width of the frame.
We can directly use the heatmap produced by the Top-Down pose estimator as the target heatmap,
which should be zero-padded to match the original frame given the corresponding bounding box.
In case we have only coordinate-triplets $(x_k, y_k, c_k)$ of skeleton joints,
we can obtain a joint heatmap $\mJ$ by composing $K$ gaussian maps centered at every joint:
\begin{equation}
	\mJ_{kij} = e^{- \frac{(i - x_k)^2 + (j - y_k)^2}{2 * \sigma^2}} * c_k, 
	\label{eq-joint-heatmap}
\end{equation}
$\sigma$ controls the variance of gaussian maps,
and $(x_k,y_k)$ and $c_k$ are respectively the location and confidence score of the $k$-th joint.
We can also create a limb heatmap $\mL$:
\begin{equation}
	\mL_{kij} = e^{- \frac{ \gD((i, j), seg[a_k, b_k])^2 }{2 * \sigma^2}} * \min(c_{a_k}, c_{b_k}). 
	\label{eq-limb-heatmap}
\end{equation}
The $k_{th}$ limb is between two joints $a_k$ and $b_k$. 
The function $\gD$ calculates the distance from the point $(i, j)$ to the segment $[(x_{a_k}, y_{a_k}), (x_{b_k}, y_{b_k})]$. 
It is worth noting that although the above process assumes a single person in every frame,
we can easily extend it to the multi-person case,
where we directly accumulate the $k$-th gaussian maps of all persons
without enlarging the heatmap.
Finally, a 3D heatmap volume is obtained by stacking all heatmaps ($\mJ$ or $\mL$) along the temporal dimension,
which thus has the size of $K\times T \times H \times W$.

In practice, we further apply two techniques to reduce the redundancy of 3D heatmap volumes.
(1) \textbf{Subjects-Centered Cropping.}
Making the heatmap as large as the frame is inefficient,
especially when the persons of interest only act in a small region.
In such cases, we first find the smallest bounding box that envelops \emph{all} the 2D poses across frames.
Then we crop all frames according to the found box and resize them to the target size.
Consequently, the size of the 3D heatmap volume can be reduced spatially while all 2D poses and their motion are kept.
(2) \textbf{Uniform Sampling.}
The 3D heatmap volume can also be reduced along the temporal dimension by sampling a subset of frames.
Unlike previous works on RGB-based action recognition,
where researchers usually sample frames in a short temporal window,
such as sampling frames in a 64-frame temporal window as in SlowFast~\cite{feichtenhofer2019slowfast},
we propose to use a uniform sampling strategy~\cite{wang2016temporal} for 3D-CNNs instead.
In particular, to sample $n$ frames from a video,
we divide the video into $n$ segments of equal length
and randomly select one frame from each segment.
The uniform sampling strategy is better at maintaining the global dynamics of the video.
Our empirical studies show that the uniform sampling strategy is significantly beneficial for skeleton-based action recognition. 
More illustration about generating 3D heatmap volumes is provided in Sec~\ref{sec-generating}.

\subsection{3D-CNN for Skeleton-based Action Recognition}
\label{sec-PoseConv3D}

For skeleton-based action recognition,
GCN has long been the mainstream backbone.
In contrast, 3D-CNN, an effective network structure commonly used in RGB-based action recognition
\cite{carreira2017quo,hara2018can,feichtenhofer2019slowfast}, is less explored in this direction.
To demonstrate the power of 3D-CNN in capturing spatiotemporal dynamics of skeleton sequences,
we design two families of 3D-CNNs, namely \textbf{PoseConv3D} for the \emph{Pose} modality and \textbf{RGBPose-Conv3D} for the \emph{RGB+Pose} dual-modality.

\begin{table}[t]
	\captionsetup{font=small, position=top}
	\caption{ 
	\textbf{Evalution of PoseConv3D variants. }
	`s' indicates shallow (fewer layers);
	`HR' indicates high-resolution (double height \& width); 
	`wd' indicates wider network with double channel size. }
	\label{tab-variant}
	\vspace{-2mm}
	\centering
	\resizebox{.9\linewidth}{!}{
	\tablestyle{6pt}{1.2}
	\begin{tabular}{ccccc}
	\shline
	Backbone  & Variant  & NTU60-XSub & FLOPs & Params \\ 
	\shline
	SlowOnly  & -		 & 93.7       & 15.9G & 2.0M   \\
	SlowOnly  & HR       & 93.6       & 73.0G & 8.0M   \\
	SlowOnly  & wd       & 93.7       & 54.9G & 7.9M   \\
	\shline
	C3D		  & -        & 93.0       & 25.2G & 6.9M   \\
	C3D		  & s        & 92.9       & 16.8G & 3.4M   \\
	\shline
	X3D		  & -        & 92.6       & 1.1G  & 531K   \\
	X3D		  & s        & 92.3       & 0.6G  & 241K   \\
	\shline
	\end{tabular}}
	\vspace{-4mm}
\end{table}

\noindent\textbf{PoseConv3D. }
PoseConv3D focuses on the modality of human skeletons,
which takes 3D heatmap volumes as input and can be instantiated with various 3D-CNN backbones.
Two modifications are needed to adapt 3D-CNNs to skeleton-based action recognition:
(1)
down-sampling operations in early stages are removed from the 3D-CNN since the spatial resolution of 3D heatmap volumes does not need to be as large as RGB clips (4\x~smaller in our setting);
(2) 
a shallower (fewer layers) and thinner (fewer channels) network is sufficient to model spatiotemporal dynamics of human skeleton sequences
since 3D heatmap volumes are already mid-level features for action recognition.
Based on these principles, we adapt three popular 3D-CNNs: C3D~\cite{tran2015learning}, SlowOnly~\cite{feichtenhofer2019slowfast},
and X3D~\cite{feichtenhofer2020x3d}, to skeleton-based action recognition
(Table~\ref{tab-PoseConv3D-arch} demonstrates the architectures of the three backbones as well as their variants).
The different variants of adapted 3D-CNNs are evaluated on the NTURGB+D-XSub benchmark (Table~\ref{tab-variant}).
Adopting a lightweight version of 3D-CNNs can significantly reduce the computational complexity
at the cost of a slight recognition performance drop ($\le 0.3\%$ for all 3D backbones).
In experiments, we use SlowOnly as the default backbone, considering its simplicity (directly inflated from ResNet)
and good recognition performance.
PoseConv3D can outperform representative GCN / 2D-CNN counterparts across various benchmarks, both in accuracy and efficiency.
More importantly, the interoperability between PoseConv3D and popular networks for RGB-based action recognition
makes it easy to involve human skeletons in multi-modality fusion.

\noindent\textbf{RGBPose-Conv3D. }
To show the interoperability of PoseConv3D,
we propose RGBPose-Conv3D for the early fusion of human skeletons and RGB frames.
It is a two-stream 3D-CNN with two pathways that respectively process RGB modality and Pose modality.
While a detailed instantiation of RGBPose-Conv3D is included in Sec~\ref{sec-rgbpose2stream},
the architecture of RGBPose-Conv3D follows several principles in general: (1) the two pathways are asymmetrical due to the different characteristics of the two modalities:
Compared to the RGB pathway,
the pose pathway has a smaller channel width, a smaller depth, as well as a smaller input spatial resolution.
(2) Inspired by 
SlowFast~\cite{feichtenhofer2019slowfast}, 
bidirectional lateral connections between the two pathways are added to promote early-stage feature fusion between two modalities.
To avoid overfitting,
RGBPose-Conv3D is trained with two individual cross-entropy losses respectively for each pathway. 
In experiments, we find that early-stage feature fusion, achieved by lateral connections, leads to consistent improvement compared to late-fusion only.

\section{Experiments}

\subsection{Dataset Preparation}
We use six datasets in our experiments: FineGYM~\cite{shao2020finegym}, NTURGB+D~\cite{Shahroudy_2016_NTURGBD, Liu_2019_NTURGBD120}, 
Kinetics400~\cite{carreira2017quo, yan2018spatial}, UCF101~\cite{soomro2012ucf101}, HMDB51~\cite{kuehne2011hmdb} 
and Volleyball~\cite{ibrahim2016hierarchical}. 
Unless otherwise specified, we use the Top-Down approach for pose extraction:
the detector is Faster-RCNN~\cite{ren2015faster} with the ResNet50 backbone, the pose estimator is 
HRNet~\cite{sun2019deep} pre-trained on COCO-keypoint~\cite{lin2014microsoft}.
For all datasets except FineGYM, 2D poses are obtained by directly applying Top-Down pose estimators to RGB inputs.
We report the \textbf{Mean Top-1} accuracy for FineGYM and \textbf{Top-1} accuracy for other datasets.
We adopt the 3D ConvNets implemented in MMAction2~\cite{2020mmaction2} in experiments. 

\noindent
\textbf{FineGYM. } 
FineGYM is a fine-grained action recognition dataset with 29K videos of 99 fine-grained gymnastic action classes. 
During pose extraction, we compare three different kinds of person bounding boxes: 
1. Person bounding boxes predicted by the detector (\textbf{Detection});
2. GT bounding boxes for the athlete in the first frame, tracking boxes for the rest frames (\textbf{Tracking}).
3. GT bounding boxes for the athlete in all frames (\textbf{GT}).
In experiments, we use human poses extracted with the third kind of bounding boxes unless otherwise noted.

\noindent
\textbf{NTURGB+D. } 
NTURGB+D is a large-scale human action recognition dataset collected in the lab.
It has two versions, namely NTU-60 and NTU-120 (a superset of NTU-60): NTU-60 contains 57K videos of 60 human actions, while NTU-120 contains 114K videos of 120 human actions.
The datasets are split in three ways: 
Cross-subject (\textbf{X-Sub}), Cross-view (\textbf{X-View}, for NTU-60), Cross-setup (\textbf{X-Set}, for NTU-120), 
for which action subjects, camera views, camera setups are different in training and validation.
The 3D skeletons collected by sensors are available for this dataset. 
Unless otherwise specified, we conduct experiments on the \textbf{X-sub} splits for NTU-60 and NTU-120.

\noindent
\textbf{Kinetics400, UCF101, and HMDB51. } 
The three datasets are general action recognition datasets collected from the web.
Kinetics400 is a large-scale video dataset with 300K videos from 400 action classes.
UCF101 and HMDB51 are smaller, contain 13K videos from 101 classes and 6.7K videos from 51 classes, respectively.
We conduct experiments using 2D-pose annotations extracted with our Top-Down pipeline. 

\noindent
\textbf{Volleyball. }
Volleyball is a group activity recognition dataset with 4830 videos of 8 group activity classes.
Each frame contains approximately 12 persons, while only the center frame has annotations for GT person boxes.
We use tracking boxes from~\cite{SendoMVA2019Heatmapping} for pose extraction.

\begin{table*}[t]
	\vspace{2mm}
	\begin{minipage}{.58\textwidth}
		\captionsetup{font=small}
		\centering
		\captionof{table}{ \textbf{PoseConv3D \textit{v.s.} GCN. } 
			We compare the performance of PoseConv3D and GCN on several datasets. 
			For PoseConv3D, we report the results of 1/10-clip testing. 
			We exclude parameters and FLOPs of the FC layer, since it depends on the number of classes.}
		\label{tab-gcnvscnn}
		\vspace{-2mm}
		\resizebox{\columnwidth}{!}{
		\tablestyle{7pt}{1.1}
			\begin{tabular}{c|ccc|cccc}
				\shline
				& \multicolumn{3}{c|}{MS-G3D} & \multicolumn{4}{c}{Pose-SlowOnly} \\ 
				\shline
				Dataset & Acc & Params & FLOPs & 1-clip & 10-clip & Params & FLOPs \\ 
				\shline
				FineGYM & 92.0  & 2.8M & 24.7G  & \textbf{92.4} & \textbf{93.2} & \multirow{4}{*}{\textbf{2.0M}} & \multirow{4}{*}{\textbf{15.9G}} \\ 
				NTU-60 & 91.9  & 2.8M &  16.7G  & \textbf{93.1} & \textbf{93.7} & & \\
				NTU-120 & 84.8  & 2.8M & 16.7G  & \textbf{85.1} & \textbf{86.0} & & \\
				Kinetics400 & \textbf{44.9}  & 2.8M & 17.5G  & 44.8 & \textbf{46.0} & & \\
				\shline
		\end{tabular}}
		\vspace{-2mm}
		\centering
		\captionof{table}{ \textbf{Recognition performance w. different dropping KP probabilities.} PoseConv3D is more robust to input perturbations. }
		\label{tab-robustness}
		\resizebox{.8\columnwidth}{!}{
			\tablestyle{8pt}{1.1}
			\begin{tabular}{c|ccccc} \shline
				Method / $p$   & 0    & 1/8  & 1/4  & 1/2  & 1    \\ 
				\shline
				MS-G3D & 92.0 & 91.0 & 90.2 & 86.5 & 77.7 \\
				+ robust training	& 90.9	& 91.0	& 91.0	& 91.0	& 90.6 \\ 
				\shline
				Pose-SlowOnly & \textbf{92.4} & \textbf{92.4} & \textbf{92.3} & \textbf{92.1} & \textbf{91.5} \\ 
				\shline
			\end{tabular}}
	\end{minipage}
	\hfill
	\begin{minipage}{.40\textwidth}
		\centering 
		\captionsetup{font=small}
		\captionof{table}{ 
				\textbf{Train/Test w. different pose annotations. } 
				PoseConv3D shows great generalization capability in the cross-PoseAnno setting
				(LQ for low-quality; HQ for high-quality). }
			\label{tab-generalization}
		\captionsetup[subfloat]{font=footnotesize,position=bottom}
		\subfloat[Train/Test w. Pose from different estimators. \label{tab-robustxanno}]{
			\resizebox{\columnwidth}{!}{
			\tablestyle{4pt}{1.2}
			\begin{tabular}{c|ccc}
				\shline
					   & \multicolumn{3}{c}{Train $\rightarrow$ Test} \\ \cline{2-4}
					   & HQ $\rightarrow$ LQ   & LQ $\rightarrow$ HQ  & LQ $\rightarrow$ LQ  \\ \shline
				MS-G3D    & 79.3  & 87.9 & 89.0  \\ \hline
				PoseConv3D & \textbf{86.5} & \textbf{91.6} & \textbf{90.7} \\ \shline
			\end{tabular}}}
		\vspace{2mm}
		\subfloat[Train/Test w. Pose extracted with different boxes. \label{tab-robustxbox}]{
			\resizebox{\columnwidth}{!}{
			\tablestyle{4pt}{1.2}
			\begin{tabular}{c|ccc}
				\shline
					   & \multicolumn{3}{c}{Train $\rightarrow$ Test} \\ \cline{2-4}
					   & HQ $\rightarrow$ LQ   & LQ $\rightarrow$ HQ  & LQ $\rightarrow$ LQ  \\ \shline
				MS-G3D    & 78.5  & 89.1 & 82.9  \\ \hline
				PoseConv3D & \textbf{82.1} & \textbf{90.6} & \textbf{85.4} \\ \shline
			\end{tabular}}}
			\vspace{-2mm}
	\end{minipage}
	\vspace{-2mm}
\end{table*}

\subsection{Good properties of PoseConv3D}
To elaborate on the good properties of 3D convolutional networks over graph networks, we compare Pose-SlowOnly with MS-G3D~\cite{liu2020disentangling}, a representative GCN-based approach in multiple dimensions.
Two models take exactly the \textbf{same} input (coordinate-triplets for GCN, heatmaps generated from coordinate-triplets for PoseConv3D). 

\noindent\textbf{Performance \& Efficiency.}
In performance comparison between PoseConv3D and GCN, 
we adopt the input shape 48\x 56\x 56 for PoseConv3D.
Table~\ref{tab-gcnvscnn} shows that under such configuration,
PoseConv3D is lighter than the GCN counterpart, both in the number of parameters and FLOPs.
Though being light-weighted, PoseConv3D achieves competitive performance on different datasets.
The 1-clip testing result is better than or comparable with a state-of-the-art GCN while requiring much less computation.
With 10-clip testing, PoseConv3D consistently outperforms the state-of-the-art GCN.
Only PoseConv3D can take advantage of multi-view testing since it subsamples the entire heatmap volumes to form each input.
Besides, PoseConv3D uses the same architecture and hyperparameters for different datasets, 
while GCN relies on heavy tuning of architectures and hyperparameters on different datasets~\cite{liu2020disentangling}.

\noindent
\textbf{Robustness.} 
To test the robustness of both models, we can drop a proportion of keypoints in the input and see how such perturbation will affect the final accuracy.
Since limb keypoints\footnote{There are eight limb keypoints: bow, wrist, knee, ankle (left/right).} 
are more critical for gymnastics than the torso or face keypoints, 
we test both models by randomly dropping one limb keypoint in each frame with probability $p$.
In Table~\ref{tab-robustness}, we see that PoseConv3D is highly robust to input perturbations:
dropping one limb keypoint per frame leads to a moderate drop (less than 1\%) in Mean-Top1,
while for GCN, it's 14.3\%. 
Someone would argue that we can train GCN with the noisy input, 
similar to the dropout operation~\cite{srivastava2014dropout}.
However, even under this setting, the Mean-Top1 accuracy of GCN still drops by 1.4\% for the case $p=1$.
Besides, with robust training, there will be an additional 1.1\% drop for the case $p=0$.
The experiment results show that PoseConv3D significantly outperforms GCN in terms of robustness for pose recognition.

\noindent
\textbf{Generalization.}
To compare the generalization of GCN and 3D-CNN, we design a cross-model check on FineGYM.
Specifically, we use two models,~\ie, HRNet (Higher-Quality, or HQ for short) and MobileNet (Lower-Quality, LQ) for pose estimation and train two PoseConv3D on top, respectively.
During testing, we feed LQ input into the model trained with HQ one and vice versa.
From Table~\ref{tab-robustxanno}, we see that
the accuracy drops less when using lower-quality poses for both training \& testing with PoseConv3D compared to GCN.
Similarly, we can also vary the source of person boxes, using either \textbf{GT} boxes (HQ) or \textbf{tracking} results (LQ) for training and testing.
The results are shown in Table~\ref{tab-robustxbox}.
The performance drop of PoseConv3D is also much smaller than GCN.

\noindent
\textbf{Scalability.}
The computation of GCN scales linearly with the increasing number of persons in the video,
making it less efficient for group activity recognition. 
We use an experiment on the Volleyball dataset~\cite{ibrahim2016hierarchical} to prove that. 
Each video in the dataset contains 13 persons and 20 frames.
For GCN, the corresponding input shape will be 13\x 20\x 17\x 3, \textbf{13} times larger than the input for one person. 
Under such configuration, the number of parameters and FLOPs for GCN is 2.8M and 7.2G (13\x).
For PoseConv3D, we can use one \textbf{single} heatmap volume (with shape 17\x 12\x 56\x 56) to represent all 13 persons\footnote{
In experiments, we find that using a single heatmap volume to represent all people is the best practice (compared to using one heatmap volume for each person). 
Please refer to Sec~\ref{sec-groupractice} for more details. }. 
The base channel-width of Pose-SlowOnly is set to 16, leading to only 0.52M parameters and 1.6 GFLOPs.
Despite the much smaller parameters and FLOPs, PoseConv3D achieves 91.3\% Top-1 accuracy on Volleyball-validation, 2.1\% higher than the GCN-based approach.

\begin{table}[t]
	\captionsetup{font=small, position=top}
	\caption{\textbf{The design of RGBPose-Conv3D. } 
		Bi-directional lateral connections outperform uni-directional ones in the early stage feature fusion.}
	\label{tab-bi-directional}
	\vspace{-2mm}
	\centering
	\resizebox{\linewidth}{!}{
	\tablestyle{4pt}{1.2}
	\begin{tabular}{c|cccc}
		\shline
		& late fusion & RGB $\rightarrow$ Pose & Pose $\rightarrow$ RGB & RGB $\leftrightarrow$ Pose \\ \shline
		1-clip  & 92.6       & 93.0    & 93.4    & \textbf{93.6}  \\ \hline
		10-clip & 93.4       & 93.7    & 93.8    & \textbf{94.1}  \\ \shline
		\end{tabular}}
		\vspace{-2mm}
\end{table}

\begin{table}[t]
	\captionsetup{font=small, position=top}
	\caption{ \textbf{The universality of RGBPose-Conv3D. } 
		The \textbf{early+late} fusion strategy works both on RGB-dominant NTU-60 and Pose-dominant FineGYM. }
	\label{tab-universality}
	\vspace{-2mm}
	\centering
	\resizebox{\linewidth}{!}{
	\tablestyle{4pt}{1.2}
	\begin{tabular}{c|cccc}
		\shline
		& RGB         & Pose        & late fusion  & early+late fusion  \\ \shline
		FineGYM & 87.2 / 88.5 & 91.0 / 92.0 & 92.6 / 93.4 & \textbf{93.6 / 94.1} \\ \hline
		NTU-60  & 94.1 / 94.9 & 92.8 / 93.2 & 95.5 / 96.0 & \textbf{96.2 / 96.5} \\ \shline
	\end{tabular}}
	\vspace{-4mm}
\end{table}

\begin{table*}[t]
	\captionsetup{font=small, position=top}
	\vspace{-3mm}
	\centering 
	\caption{ \textbf{PoseConv3D is better or comparable to previous state-of-the-arts. } 
		With estimated high-quality 2D skeletons and the great capacity of 3D-CNN to learn spatiotemporal features, 
		PoseConv3D achieves superior performance across \textbf{5 out of 6} benchmarks.
		$\mJ, \mL$ means using joint/limb-based heatmaps.
		++ denotes using the same human skeletons as ours.
		Numbers with * are reported by \cite{shao2020finegym}.
	} 
	\label{tab-posesota}
	\vspace{-2mm}
	\resizebox{.95\linewidth}{!}{
	\tablestyle{6pt}{1.2}
	\begin{tabular}{c|cccccc}
	\shline
	Method       						& NTU60-XSub & NTU60-XView & NTU120-XSub & NTU120-XSet & Kinetics & FineGYM \\ \shline
	ST-GCN~\cite{yan2018spatial} 		& 81.5       & 88.3        & 70.7        & 73.2        & 30.7     & 25.2*   \\ 
	AS-GCN~\cite{li2019actional}   	& 86.8       & 94.2        & 78.3        & 79.8        & 34.8     & -       \\ 
	RA-GCN~\cite{song2020richly} 		& 87.3       & 93.6        & 81.1        & 82.7        & -        & -       \\ 
	AGCN~\cite{shi2019two}         	& 88.5       & 95.1        & -           & -           & 36.1     & -       \\ 
	DGNN~\cite{shi2019skeleton} 	  	& 89.9       & 96.1        & -           & -           & 36.9     & -       \\ 
	FGCN~\cite{yang2020feedback}       & 90.2       & 96.3        & 85.4        & 87.4        & -        & -       \\ 
	Shift-GCN~\cite{cheng2020skeleton}	& 90.7       & 96.5        & 85.9        & 87.6        & -        & -       \\ 
	DSTA-Net~\cite{shi2020decoupled}   & 91.5       & 96.4        & 86.6        & 89.0        & -        & -       \\ 
	MS-G3D~\cite{liu2020disentangling} & 91.5       & 96.2        & 86.9        & 88.4        & 38.0     & -       \\ \shline
	MS-G3D ++							& 92.2       & 96.6        & \textbf{87.2}  & 89.0     & 45.1     & 92.6    \\ \shline
	PoseConv3D ($\mJ$)      			& \textbf{93.7}   & \textbf{96.6}  & 86.0   & \textbf{89.6} & \textbf{46.0} & \textbf{93.2}	\\ \shline
	PoseConv3D ($\mJ+\mL$)    			& \textbf{94.1}   & \textbf{97.1}  & 86.9   & \textbf{90.3} & \textbf{47.7} & \textbf{94.3} \\ \shline
	\end{tabular}}
	\vspace{-2mm}
\end{table*}

\begin{table*}[t]
	\captionsetup{font=small}
	\centering 
	\caption{ \textbf{Comparison to the state-of-the-art of Multi-Modality Action Recognition. } 
			 Strong recognition performance is achieved on multiple benchmarks with multi-modality fusion. 
			 R, F, P indicate RGB, Flow, Pose. } 
	\label{tab-rgbposesota}
	\vspace{-2mm}
	\captionsetup[subfloat]{font=small}
	\captionsetup[subffloat]{justification=centering}
	\resizebox{\linewidth}{!}{
	\subfloat[\label{tab-rgbpose-conv3d} Mulit-modality action recognition with \textit{RGBPose-Conv3D}. ]{
		\tablestyle{6pt}{1.3}
		\begin{tabular}{c|c|c}
			\shline
			Dataset	& Previous state-of-the-art	& \textbf{Ours}  \\ \shline
			FineGYM-99        		& \begin{tabular}[c]{@{}c@{}} 87.7 (R) \cite{kwon2021learning} \end{tabular} & \textbf{95.6} (R + P) \\ \hline
			NTU60 (X-Sub / X-View)  & \begin{tabular}[c]{@{}c@{}} 95.7 / 98.9 (R + P) \cite{davoodikakhki2020hierarchical} \end{tabular} & \textbf{97.0 / 99.6} (R + P) \\ \hline
			NTU120 (X-Sub / X-Set)  & \begin{tabular}[c]{@{}c@{}} 90.7 / 92.5 (R + P) \cite{das2021vpn++} \end{tabular} & \textbf{95.3 / 96.4} (R + P) \\ \shline
		\end{tabular}}
	\hspace{1mm}
	\subfloat[\label{tab-rgbpose-latefusion} Mulit-modality action recognition with \textit{LateFusion}.\footnotemark ]{
		\tablestyle{6pt}{1.3}
		\begin{tabular}{c|c|c|c}
			\shline
			Dataset  & Previous state-of-the-art & \textbf{Ours (Pose)} & \textbf{Ours (Fused)}  \\ \shline
			Kinetics400	& \begin{tabular}[c]{@{}c@{}} 84.9 (R) \cite{liu2021video} \end{tabular} & 47.7 & \textbf{85.5 } (R + P)\\ \hline
			UCF101 & \begin{tabular}[c]{@{}c@{}} 98.6 (R + F) \cite{duan2020omni} \end{tabular} & 87.0 & \textbf{98.8} (R + F + P) \\ \hline
			HMDB51 	& \begin{tabular}[c]{@{}c@{}} 83.8 (R + F)	\cite{duan2020omni} \end{tabular} & 69.3 & \textbf{85.0} (R + F + P) \\ \shline
		\end{tabular}}}
	\vspace{-4mm}
\end{table*}

\subsection{Multi-Modality Fusion with RGBPose-Conv3D}
The 3D-CNN architecture of PoseConv3D makes it more flexible to fuse pose with other modalities via some early fusion strategies.
For example, in \emph{RGBPose}-Conv3D, lateral connections between the \emph{RGB}-pathway and \emph{Pose}-pathway are exploited for cross-modality feature fusion in the early stage.
In practice, we first train two models for RGB and Pose modalities separately and use them to initialize the \emph{RGBPose}-Conv3D.
We continue to finetune the network for several epochs to train the lateral connections. 
The final prediction is achieved by late fusing the prediction scores from both pathways.
\emph{RGBPose}-Conv3D can achieve better fusing results with \textbf{early+late} fusion.

We first compare uni-directional lateral connections and bi-directional lateral connections in Table~\ref{tab-bi-directional}.
The result shows that bi-directional feature fusion is better than uni-directional ones for RGB and Pose. 
With bi-directional feature fusion in the early stage, the \textbf{early+late} fusion with 1-clip testing can outperform the \textbf{late} fusion with 10-clip testing. 
Besides, \emph{RGBPose}-Conv3D also works in situations when the importance of two modalities is different.
The pose modality is more important in FineGYM and vice versa in NTU-60. 
Yet we observe performance improvement by \textbf{early+late} fusion on both of them in Table~\ref{tab-universality}. 
We demonstrate the detailed instantiation of \emph{RGBPose}-Conv3D we used in Sec~\ref{sec-rgbpose2stream}. 

\subsection{Comparisons with the state-of-the-art}

\noindent\textbf{Skeleton-based Action Recognition. }
In Table~\ref{tab-posesota}, we compare PoseConv3D with prior works for skeleton-based action recognition. 
Prior works (Table~\ref{tab-posesota} upper) use 3D skeletons collected with Kinect for NTURGB+D, 
2D skeletons extracted with OpenPose for Kinetics 
(details for FineGYM skeleton data are unknown).
PoseConv3D adopts 2D skeletons extracted with good practices introduced in Sec~\ref{sec-pose-extraction}, which have better quality. 
We instantiate PoseConv3D with the SlowOnly backbone, feed 3D heatmap volumes of shape 48\x 56\x 56 as inputs, and report the accuracy obtained by 10-clip testing. 
For a fair comparison, we also evaluate the state-of-the-art MS-G3D with our 2D human skeletons (\emph{MS-G3D++}):
\emph{MS-G3D++} directly takes the extracted coordinate-triplets $(x, y, c)$ as inputs, while \emph{PoseConv3D} takes pseudo heatmaps generated from the coordinate-triplets as inputs. 
With high quality 2D human skeletons, \emph{MS-G3D++} and \emph{PoseConv3D} both achieve far better performance than previous state-of-the-arts, 
demonstrating the \textbf{importance} of the proposed practices for pose extraction in skeleton-based action recognition. 
When both take high-quality 2D poses as inputs, PoseConv3D outperforms the state-of-the-art MS-G3D across \textbf{5 of 6} benchmarks, 
showing its great spatiotemporal feature learning capability. 
PoseConv3D achieves by far the best results on \textbf{3 of 4} NTURGB+D benchmarks. 
On Kinetics, PoseConv3D surpasses MS-G3D++ by a noticeable margin, significantly outperforming all previous methods. 
Except for the baseline reported in \cite{shao2020finegym}, no work aims at skeleton-based action recognition on FineGYM before,
while our work first improves the performance to a decent level.

\footnotetext{For K400, we fuse PoseConv3D Pose predictions (Top1 acc 47.7\%) with VideoSwin~\cite{liu2021video} RGB predictions.
For UCF101 and HMDB51, we fuse PoseConv3D Pose predictions (with K400 PoseConv3D pre-training, 
87\% Top1 acc on UCF101, 69.3\% Top1 acc on HMDB51, details in Sec~\ref{sec-mmar-ucfhmdb}) 
with OmniSource~\cite{duan2020omni} RGB+Flow predictions.}

\noindent
\textbf{Multi-modality Fusion. }
As a powerful representation itself, skeletons are also complementary to other modalities, like RGB appearance. 
With multi-modality fusion (\emph{RGBPose-Conv3D} or \emph{LateFusion}),
we achieve state-of-the-art results across \textbf{8} different video recognition benchmarks. 
We apply the proposed \emph{RGBPose-Conv3D} to FineGYM and 4 NTURGB+D benchmarks, 
using R50 as the backbone; 16, 48 as the temporal length for \emph{RGB}/\emph{Pose}-Pathway. 
Table~\ref{tab-rgbpose-conv3d} shows that our \textbf{early+late} fusion achieves excellent performance across various benchmarks.
We also try to fuse the predictions of PoseConv3D directly with other modalities with \emph{LateFusion}. 
Table~\ref{tab-rgbpose-latefusion} shows that late fusion with the Pose modality can push the recognition precision to a new level.
We achieve the new state-of-the-art on three action recognition benchmarks: Kinetics400, UCF101, and HMDB51.
On the challenging Kinetics400 benchmark, 
fusing with PoseConv3D predictions increases the recognition accuracy by 0.6\% beyond the state-of-the-art~\cite{liu2021video},
which is strong evidence for the complementarity of the Pose modality.

\subsection{Ablation on Heatmap Processing}

\noindent\textbf{Subjects-Centered Cropping. }
Since the sizes and locations of persons can vary a lot in a dataset, 
focusing on the action subjects is the key to reserving as much information as possible with a relatively small $H\times W$ budget.
To validate this, we conduct a pair of experiments on FineGYM with input size 32\x 56\x 56, with or without subjects-centered cropping.
We find that subjects-centered cropping is helpful in data preprocessing, which improves the Mean-Top1 by 1.0\%, from 91.7\% to 92.7\%.

\begin{figure}[t]
	\captionsetup{font=small}
	\includegraphics[width=\linewidth]{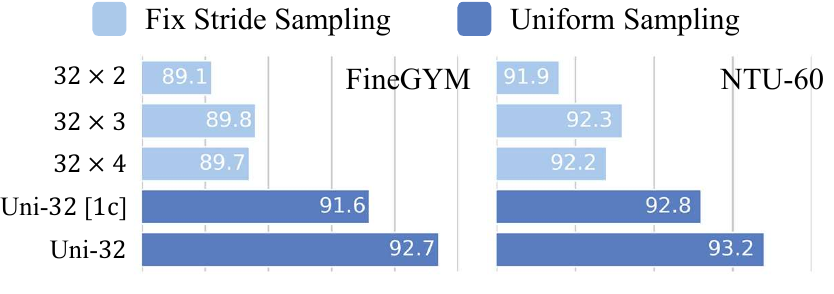} 
	\vspace{-8mm}
	\caption{ \textbf{Uniform Sampling outperforms Fix-Stride Sampling. } All results are for 10-clip testing, except Uni-32[1c], which uses 1-clip testing. }
	\label{fig-unisample}
	\vspace{-3mm}
\end{figure}

\begin{figure}[t]
	\captionsetup{font=small}
	\centering
	\includegraphics[width=\linewidth]{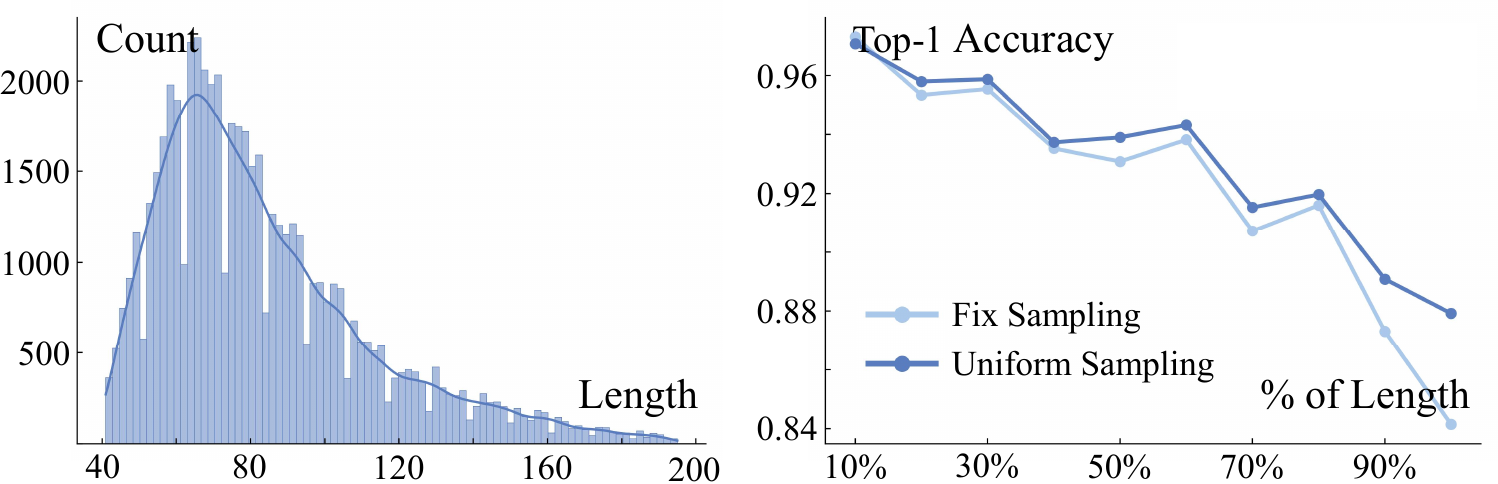} 
	\vspace{-5mm}
	\caption{ \textbf{Uniform Sampling helps in modeling longer videos. }  
		L: The length distribution of NTU60-XSub val videos.
		R: Uniform Sampling improves the recognition accuracy of longer videos.}
	\label{fig-unideeper}
	\vspace{-3mm}
\end{figure}

\noindent
\textbf{Uniform Sampling. }
The input sampled from a small temporal window may not capture the entire dynamic of the human action. 
To validate this, we conduct experiments on FineGYM and NTU-60.
For fixed stride sampling, which samples from a fixed temporal window, we try to sample 32 frames with the temporal stride 2, 3, 4; 
for uniform sampling, we sample 32 frames uniformly from the entire clip.
In testing, we adopt a fixed random seed when sampling frames from each clip to make sure the test results are reproducible. 
From Figure~\ref{fig-unisample}, we see that uniform sampling consistently outperforms sampling with fixed temporal strides.
With uniform sampling, 1-clip testing can even achieve better results than fixed stride sampling with 10-clip testing.
Note that the video length can vary a lot in NTU-60 and FineGYM.
In a more detailed analysis, we find that uniform sampling mainly improves the recognition performance 
for longer videos in the dataset (Figure~\ref{fig-unideeper}).
Besides, uniform sampling also outperforms fixed stride sampling on RGB-based recognition on the two datasets\footnote{
	Please refer to Sec~\ref{sec-uniformrgb} for details and discussions. }.

\noindent
\textbf{Pseudo Heatmaps for Joints and Limbs. }
GCN approaches for skeleton-based action recognition usually ensemble results of multiple streams (joint, bone, \textit{etc.}) 
to obtain better recognition performance~\cite{shi2019two}. 
The practice is also feasible for PoseConv3D.
Based on the coordinates $(x, y, c)$ we saved, 
we can generate pseudo heatmaps for joints and limbs. 
In general, we find that both joint heatmaps and limb heatmaps are good inputs for 3D-CNNs. 
Ensembling the results from joint-PoseConv3D and limb-PoseConv3D (namely PoseConv3D ($\mJ+\mL$)) can lead to noticeable and consistent performance improvement.

\noindent
\textbf{3D Heatmap Volumes \emph{v.s} 2D Heatmap Aggregations. }
The 3D heatmap volume is a more `lossless' 2D-pose representation,
compared to 2D pseudo images aggregating heatmaps
with colorization or temporal convolutions.
PoTion~\cite{choutas2018potion} and PA3D~\cite{yan2019pa3d} are not evaluated on popular benchmarks for skeleton-based action recognition, 
and there are no public implementations.  
In the preliminary study, we find that the accuracy of PoTion is much inferior ($\le 85\%$) to GCN or PoseConv3D (all $\ge 90\%$).
For an apple-to-apple comparison, we also re-implement PoTion, PA3D 
(with higher accuracy than reported) 
and evaluate them on UCF101, HMDB51, and NTURGB+D. 
PoseConv3D achieves much better recognition results with 3D heatmap volumes, 
than 2D-CNNs with 2D heatmap aggregations as inputs. 
With the lightweight X3D, PoseConv3D significantly outperforms 2D-CNNs, 
with comparable FLOPs and far fewer parameters (Table~\ref{tab-volumevsaggr}).

\begin{table}[t]
	\captionsetup{font=small, position=top}
	\centering 
	\caption{ \textbf{An apple-to-apple comparison between 3D heatmap volumes and 2D heatmap aggregations. }}
	\label{tab-volumevsaggr}
	\vspace{-2mm}
	\resizebox{\linewidth}{!}{
	\tablestyle{6pt}{1.3}
	\begin{tabular}{c|ccc|cc}
	\shline
	Method		 	 & HMDB51 & UCF101 & NTU60-XSub & FLOPs & Params \\ 
	\shline
	PoTion~\cite{choutas2018potion}   & 51.7   & 67.2   & 87.8   & 0.60G & 4.75M  \\
	PA3D~\cite{yan2019pa3d} & 53.5   & 69.1   & 88.6   & 0.65G  & 4.81M  \\ 
	\shline
	Pose-SlowOnly (Ours) & \textbf{58.6}   & \textbf{79.1}   & \textbf{93.7}   & 15.9G  & \textbf{2.0M} \\
	Pose-X3D-s (Ours)  & \textbf{55.6}   & \textbf{76.7}  & \textbf{92.3}   & \textbf{0.60G}  & \textbf{0.24M} \\ 
	\shline
	\end{tabular}}
	\vspace{-5mm}
\end{table}
\section{Conclusion}

In this work, we propose \textbf{PoseConv3D}:
a 3D-CNN-based approach for skeleton-based action recognition,
which takes 3D heatmap volumes as input.
PoseConv3D resolves the limitations of GCN-based approaches in \emph{robustness}, \emph{interoperability}, and \emph{scalability}.
With light-weighted 3D-ConvNets and compact 3D heatmap volumes as input,
PoseConv3D outperforms GCN-based approaches in both accuracy and efficiency.
Based on PoseConv3D, we achieve state-of-the-art on both
skeleton-based and multi-modality-based action recognition
across multiple benchmarks.

\noindent
\textbf{Acknowledgement. }
This study is supported by the General Research Funds of Hong Kong (No. 14203518),
the RIE2020 Industry Alignment Fund–Industry Collaboration Projects (IAF-ICP) Funding Initiative, 
and Shanghai Committee of Science and Technology (No. 20DZ1100800).

\appendix

\section{Visualization}
\label{sec-vis}

We provide more visualization of the extracted pose of the four datasets:
FineGYM, NTURGB+D, Kinetics400, Volleyball
to demonstrate the performance of the proposed pose extraction approach qualitatively.
You can watch the corresponding videos at \url{https://youtu.be/oS7fX9Eg2ws}. 

\begin{figure}[t]
	\captionsetup{font=small}
	\centering
	\includegraphics[width=.8\linewidth]{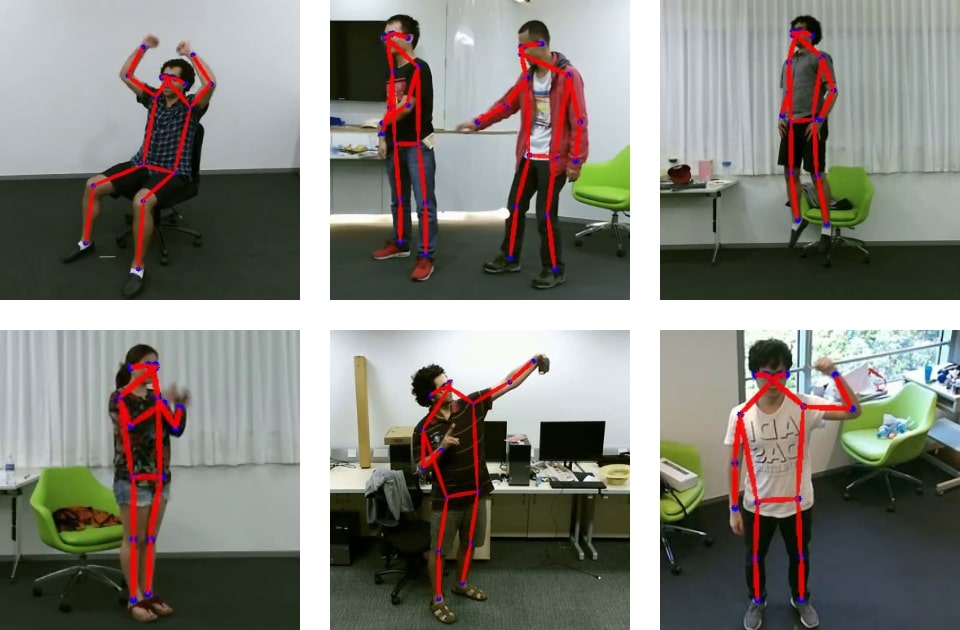}
	\vspace{-2mm}
	\caption{ \textbf{The extracted skeletons of the NTURGB+D dataset. } 
		The actions of the visualized frames are: ``cheer up'', ``touch other person's pocket'', ``jump up'', ``put the palms together'', ``taking a selfie'', ``shake fist''. }
	\vspace{-2mm}
	\label{fig-ntuvis}

\end{figure}

\textbf{NTURGB+D~\cite{Shahroudy_2016_NTURGBD, Liu_2019_NTURGBD120}. }  
Figure~\ref{fig-ntuvis} displays some examples of extracted skeletons of NTURGB+D.
Our pose extractor achieves almost perfect performance on NTURGB+D due to the simple scenarios:
the background scene is not complicated, while there are two persons at most in each frame, with little occlusion.

\begin{figure}[t]
	\captionsetup{font=small}
	\centering
	\includegraphics[width=.8\linewidth]{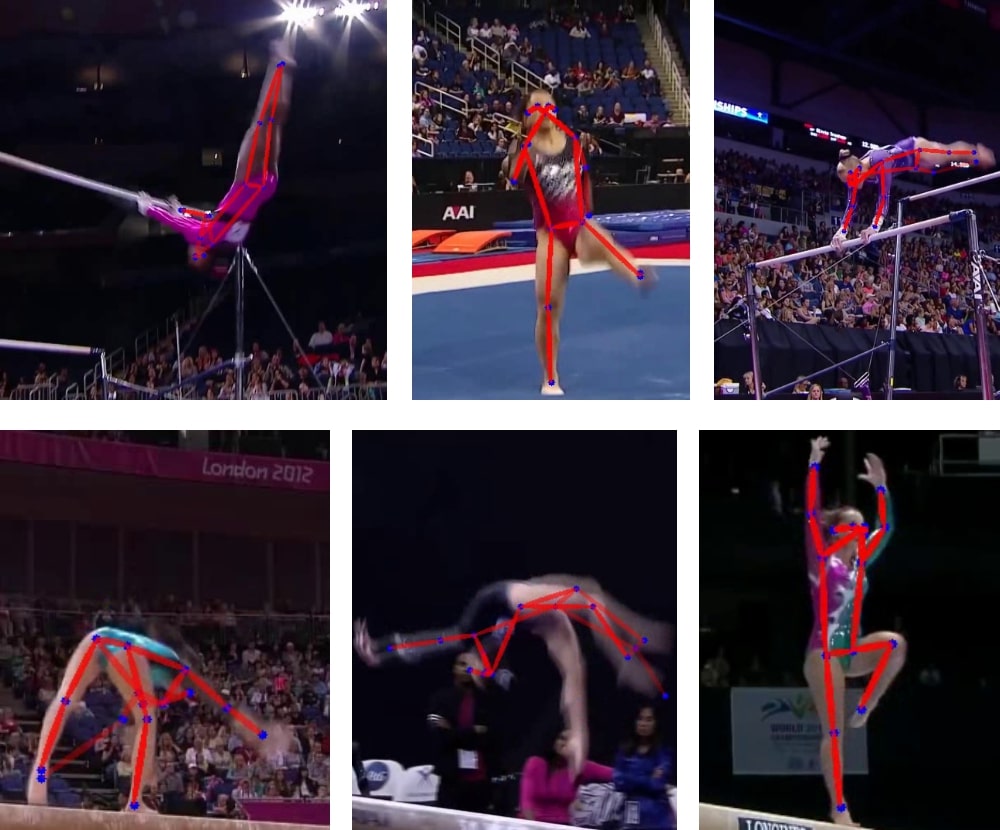}
	\vspace{-2mm}
	\caption{ \textbf{The extracted skeletons of the FineGYM dataset. } 
		The extracted skeletons are far from perfect, 
		but discriminative enough for action recognition. }
	\vspace{-4mm}
	\label{fig-gymvis}
\end{figure}

\textbf{FineGYM~\cite{shao2020finegym}. }
Figure~\ref{fig-gymvis} displays some examples of extracted skeletons of FineGYM.
Although we perform pose extraction with ground-truth bounding boxes of the athletes,
the extracted 2D poses are far from perfect.
The pose extractor is extremely easy to make mistakes for poses the rarely occur in COCO-keypoint~\cite{lin2014microsoft} or when motion blur occurs.
Even though the quality of extracted skeletons are not satisfying, 
they are still discriminative enough for skeleton-based action recognition.

\begin{figure}[t]
	\captionsetup{font=small}
	\centering
	\includegraphics[width=.8\linewidth]{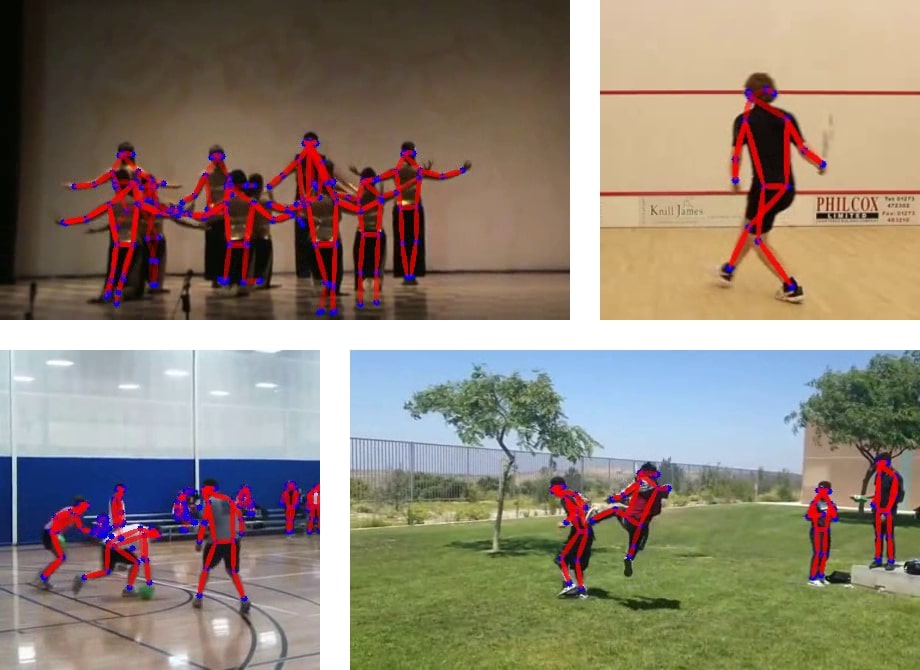}
	\vspace{-2mm}
	\caption{ \textbf{The extracted skeletons of the Kinetics400 dataset. }}
	\vspace{-2mm}
	\label{fig-kineticsvis}
\end{figure}

\textbf{Kinetics400~\cite{carreira2017quo}. }
Kinetics400 is not a human-centric dataset for action recognition.
In Kinetics videos, the person locations, scales, and the number of persons may vary a lot, 
which makes extracting human skeletons of Kinetics400 much more difficult than NTURGB+D or FineGYM.
In Figure~\ref{fig-kineticsvis}, we provide some examples that our pose estimator accurately predicts the human skeletons. 
We also discuss some failure cases in Sec~\ref{sec-error}.

\begin{figure}[t]
	\captionsetup{font=small}
	\centering
	\includegraphics[width=.8\linewidth]{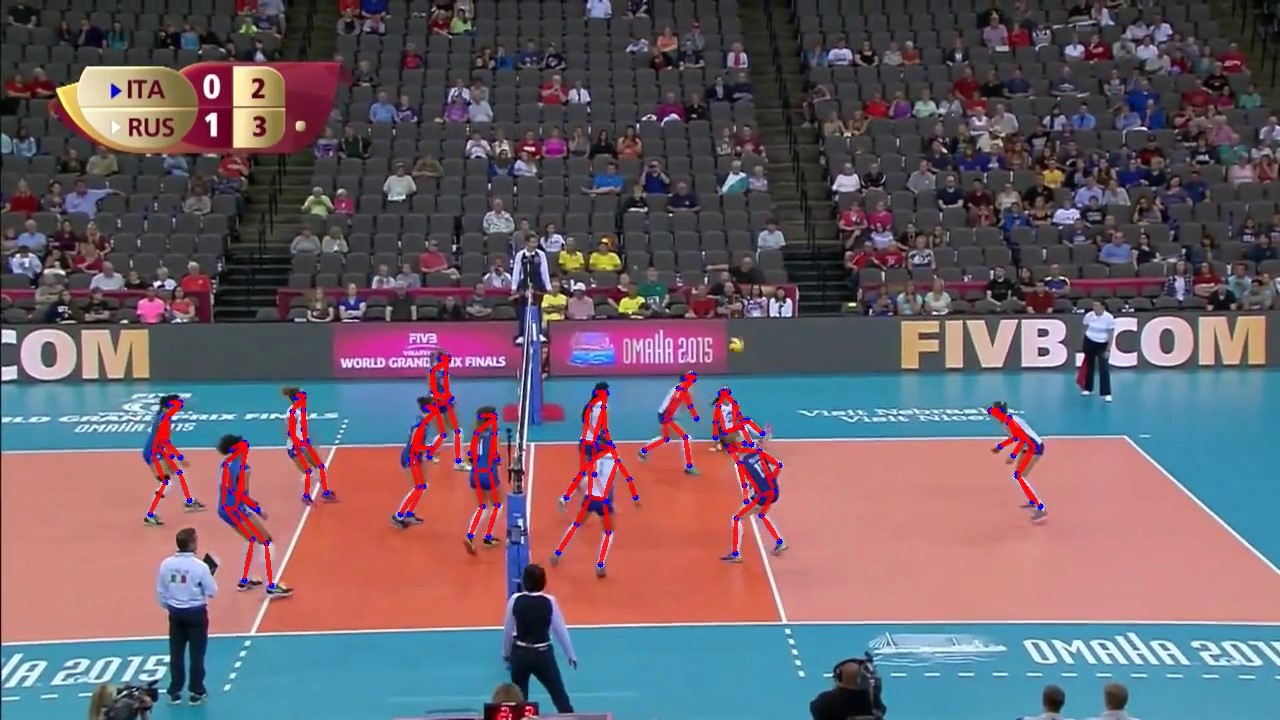}
	\vspace{-2mm}
	\caption{ \textbf{The extracted skeletons of the Volleyball dataset. }}
	\vspace{-4mm}
	\label{fig-volleyball}
\end{figure}

\textbf{Volleyball~\cite{ibrahim2016hierarchical}. } 
Volleyball is a group activity recognition dataset.
Each frame of a video contains around a dozen people (six for each team).
Most of the human poses in a volleyball video are regular ones (unlike FineGYM).
In Figure~\ref{fig-volleyball}, we see that our pose extractor can predict the human pose of each person accurately.

\section{Generating Pseudo Heatmap Volumes. }
\label{sec-generating}
In this section, we illustrate how we generate the pseudo heatmap volumes,
the input of PoseConv3D. 
We also provide a jupyter notebook named \texttt{GenPseudoHeatmaps.ipynb} in supplementary materials,
which can extract skeleton keypoints from RGB videos (optional)
and generate pseudo heatmaps based on the skeleton keypoints. 

\begin{figure*}[t]
    \captionsetup{font=small}
	\centering
	\includegraphics[width=.8\linewidth]{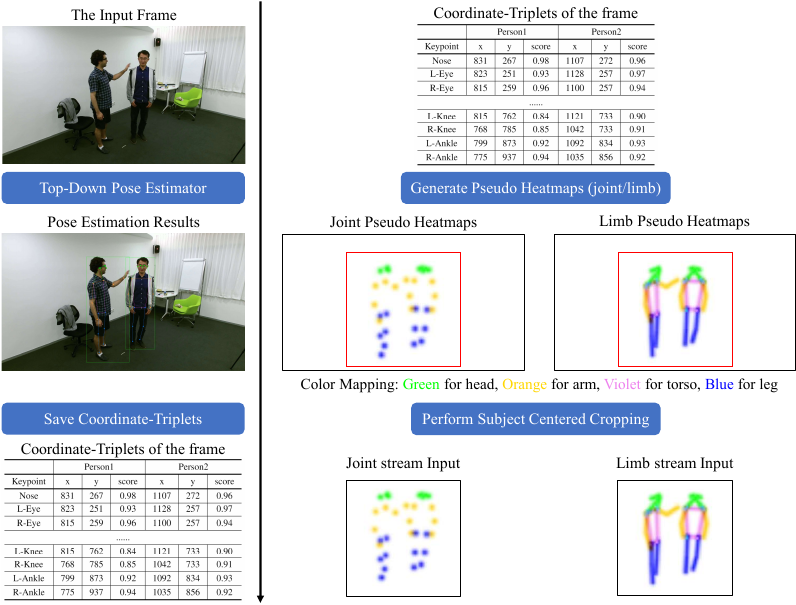}
    \vspace{-2mm}
	\caption{\textbf{The pipeline of generating the input of PoseConv3D. } 
		\textbf{Left, Pose Extraction: } We perform Top-Down pose estimation for each single frame.
                                        The estimated 2D poses are saved as coordinate-triplets: (x, y, score).
        \textbf{Right, Generating Pseudo Heatmap Volumes: } Based on the coordinate-triplets, we generate pseudo heatmaps 
                                        for joints and limbs using Eq~\ref{eq-joint-heatmap},\ref{eq-limb-heatmap}.
                                        We perform subjects-centered cropping and uniform sampling 
                                        to make the heatmap volumes compact. }
    \vspace{-5mm}
	\label{fig-datapipeline}
\end{figure*}

Figure~\ref{fig-datapipeline} illustrates the pipeline of pose extraction (RGB video $\rightarrow$ coordinate-triplets) 
and generating pseudo heatmap volumes (coordinate-triplets $\rightarrow$ 3D heatmap volumes). 
We only visualize one frame in Figure~\ref{fig-datapipeline}, while you can find the processing for an entire video 
in the notebook. 
Though a heatmap is of $K$ channels ($K=17$ for COCO-keypoints), 
we visualize it in one 2D image with color encoding. 
For pose extraction,
we use a Top-Down pose estimator instantiated with HRNet~\cite{sun2019deep} to extract the 2D poses for each person in each frame, 
and save the coordinate-triplets: (x, y, score). 
For generating pseudo heatmaps, we first perform uniform sampling, 
which will sample $T$ frames uniformly from the video and discard the remaining frames. 
We then find a global cropping box (The \red{red} box in Figure~\ref{fig-datapipeline}, same for all $T$ frames) that envelops all persons in the video, 
and crop all $T$ frames with that box to reduce the spatial size. 
You can run the entire pipeline in \texttt{GenPseudoHeatmaps.ipynb}.

\section{Detailed Architectures of PoseConv3D }

\begin{table*}[t]
    \captionsetup{font=small}
	\centering
	\caption{\textbf{PoseConv3D instantiated with: C3D, X3D, SlowOnly. }
		The dimensions of kernels are denoted by ${T\times S^2, C}$ for temporal, spatial, channel sizes.
		Strides are denoted with ${T, S^2}$ for temporal and spatial strides.
		GAP denotes global average pooling.}
    \label{tab-PoseConv3D-arch}
	\vspace{-2mm}
    \resizebox{\linewidth}{!}{
    \tablestyle{6pt}{1.5}
    \begin{tabular}{c|c|c|c|c|c|c|c}
        \shline
        stage        & C3D-s  & C3D   & X3D-s  & X3D &  SlowOnly & SlowOnly-wd & SlowOnly-HR  \\ \shline
        data layer             & \multicolumn{6}{c|}{Uniform 48, 56 \x 56}   & Uniform 48, 112 \x 112 \\ \hline
        stem layer             & \multicolumn{2}{c|}{conv 3\x 3\sqr, 32}  & \multicolumn{2}{c|}{\begin{tabular}[c]{@{}c@{}}conv 1\x 3\sqr, 24\\ stride 1, 2\sqr \\ conv 5\x 1\sqr, 24\end{tabular}} & conv 1\x 7\sqr, 32 & conv 1\x 7\sqr, 64 & conv 1\x 7\sqr, 32 \\ \hline
        stage1                 & \multicolumn{2}{c|}{\begin{tabular}[c]{@{}c@{}}maxpool 1\x 2\sqr \\ \blockctd{64}{1} \end{tabular}}  & \blockxtd{24}{54}{2}  &  \blockxtd{24}{54}{5}   &  \multicolumn{2}{c|}{None} &  \myblocks{128}{32}{3}  \\ \hline
        stage2                 & \multicolumn{2}{c|}{\begin{tabular}[c]{@{}c@{}}maxpool 1\x 2\sqr \\ \blockctd{128}{2} \end{tabular}} & \blockxtd{48}{108}{5}  &  \blockxtd{48}{108}{11}   &  \myblocks{128}{32}{4} &  \multicolumn{2}{c}{\myblocks{256}{64}{4}} \\ \hline
        stage3                 & \multicolumn{2}{c|}{\begin{tabular}[c]{@{}c@{}}maxpool 1\x 2\sqr \\ \blockctd{256}{2} \end{tabular}} & \blockxtd{96}{216}{3}  &  \blockxtd{96}{216}{7}   &  \myblockt{256}{64}{6}  &  \multicolumn{2}{c}{\myblockt{512}{128}{6}} \\ \hline
        stage4                 & None   & \blockctd{256}{2}   &   \multicolumn{2}{c|}{conv 1\x 1\sqr, 216}   &  \myblockt{512}{128}{3}  &  \multicolumn{2}{c}{\myblockt{1024}{256}{3}} \\ \hline
        \multicolumn{1}{l|}{} & \multicolumn{7}{c}{GAP, fc}                                                                                                                                                                                                                                                    \\ \hline
    \end{tabular}}
	\vspace{-3mm}
\end{table*}

\subsection{Different variants of PoseConv3D. }
In Table~\ref{tab-PoseConv3D-arch}, we demonstrate the architectures of 
the three backbones we adapted from RGB-based action recognition as well as their variants: 

\textbf{C3D~\cite{tran2015learning}. }
C3D is one of the earliest 3D-CNN developed for RGB-based action recognition (like AlexNet~\cite{krizhevsky2012imagenet} for image recognition), 
which consists of eight 3D convolution layers. 
To adapt C3D for skeleton-based action recognition,
we reduce its channel-width to half ($64\rightarrow 32$) for better efficiency. 
In addition, for Pose-C3D-s, we remove the last two convolution layers.

\textbf{X3D~\cite{feichtenhofer2020x3d}. }
X3D is a recent state-of-the-art 3D-CNN for action recognition. Replacing vanilla convolutions with depth-wise convolutions, 
X3D achieves competitive recognition performance with tiny amounts of parameters and FLOPs. 
The architecture of the adapted Pose-X3D is almost unchanged compared to the original X3D-S, except that we remove the original first stage. 
For Pose-X3D-s, we remove convolution layers from each stage uniformly by changing the hyper-parameter $\gamma_d$ from 2.2 to 1. 

\textbf{SlowOnly~\cite{feichtenhofer2019slowfast}. }
SlowOnly is a popular 3D-CNN used for RGB-based action recognition. 
It is obtained by inflating the ResNet layers in the last two stages from 2D to 3D.
To adapt SlowOnly for skeleton-based action recognition,
we reduce its channel-width to half ($64\rightarrow 32$) as well as remove the original first stage in the network. 
We also have conducted experiments with Pose-SlowOnly-wd (with channel-width 64) and Pose-SlowOnly-HR (with 2x larger input and deeper network). 
There is no performance improvement despite the much heavier backbone. 

\subsection{RGBPose-Conv3D instantiated with SlowOnly. }
\label{sec-rgbpose2stream}
RGBPose-Conv3D is a general framework for RGB-Pose dual-modality action recognition,
which can be instantiated with various 3D-CNN backbones. 
In this work, we instantiate both pathways with the SlowOnly network. 
As shown in Table~\ref{tab-rgbpose-arch}, 
the RGB pathway has a smaller frame rate and a larger channel width since RGB frames are low-level features. 
On the contrary, the Pose pathway has a larger frame rate and a smaller channel width.
Time stride convolutions are used as bi-directional lateral connections between the two pathways (after res$_3$ and res$_4$)
so that semantics of different modalities can sufficiently interact.
Besides lateral connections, the predictions of two pathways are also combined in a late fusion manner, 
which leads to further improvements in our empirical study. 
RGBPose-Conv3D is trained with two individual losses respectively for each pathway, 
as a single loss that jointly learns from two modalities leads to severe overfitting.

\begin{table}[t]
	\captionsetup{font=small}
	\centering
	\caption{\textbf{RGBPose-Conv3D instantiated with the SlowOnly backbone. }
		The dimensions of kernels are denoted by ${T\times S^2, C}$ for temporal, spatial, channel sizes.
		Strides are denoted with ${T, S^2}$ for temporal and spatial strides.
		The backbone we use is ResNet50.
		GAP denotes global average pooling.}
	\label{tab-rgbpose-arch}
	\vspace{-2mm}
	\resizebox{\columnwidth}{!}{
		\tablestyle{6pt}{1.4}
		\begin{tabular}{c|c|c|c}
			\hline
			stage & \emph{RGB} Pathway & \emph{Pose} Pathway & output sizes T\x S\sqr \\ \shline
			data layer & uniform 8,1\sqr & uniform 32,4\sqr & \begin{tabular}[c]{@{}c@{}}\emph{RGB}: 8\x 224\sqr \\ \emph{Pose}: 32\x 56\sqr \end{tabular} \\ \hline
			stem layer & \begin{tabular}[c]{@{}c@{}}conv 1\x 7\sqr, 64\\ stride 1, 2\sqr\\ maxpool 1\x 3\sqr\\ stride 1, 2\sqr\end{tabular} & \begin{tabular}[c]{@{}c@{}}conv 1\x 7\sqr, 32 \\ stride 1, 1\sqr \end{tabular} & \begin{tabular}[c]{@{}c@{}}\emph{RGB}: 8\x 56\sqr\\ \emph{Pose}: 32\x 56\sqr \end{tabular} \\ \hline
			\multirow{3}{*}{res$_2$} & \blocks{256}{64}{3}  &\multirow{3}{*}{N.A.} & \multirow{3}{*}{\begin{tabular}[c]{@{}c@{}}\emph{RGB}: 8\x 56\sqr \\ \emph{Pose}: 32\x 56\sqr \end{tabular}} \\
			& &  & \\
			& &  & \\ \hline
			\multirow{3}{*}{res$_3$} & \blocks{512}{128}{4}  & \blocks{128}{32}{4}   & \multirow{3}{*}{\begin{tabular}[c]{@{}c@{}}\emph{RGB}: 8\x 28\sqr \\ \emph{Pose}: 32\x 28\sqr \end{tabular}} \\
			& &  & \\
			& &  & \\ \hline
			\multirow{3}{*}{res$_4$} & \blockt{1024}{256}{6}  & \blockt{256}{64}{6}   & \multirow{3}{*}{\begin{tabular}[c]{@{}c@{}}\emph{RGB}: 8\x 14\sqr \\ \emph{Pose}: 32\x 14\sqr \end{tabular}} \\
			& &  & \\
			& &  & \\ \hline
			\multirow{3}{*}{res$_5$} & \blockt{2048}{512}{3}  & \blockt{512}{128}{3}   & \multirow{3}{*}{\begin{tabular}[c]{@{}c@{}}\emph{RGB}: 8\x 7\sqr \\ \emph{Pose}: 32\x 7\sqr \end{tabular}} \\
			& &  & \\
			& &  & \\ \hline
			& GAP, fc & GAP, fc  & \# classes \\ \hline
	\end{tabular}}
	\vspace{-4mm}
\end{table}
\section{Supplementary Experiments}

\begin{table*}[t]
	\centering
	\captionsetup{font=small}
	\captionsetup[subfloat]{font=footnotesize, position=bottom}
	\caption{\small \textbf{Ablation study on Pose Extraction.}}
	\vspace{-2mm}
	\label{tab-poseext}
	\resizebox{.65\linewidth}{!}{
	\subfloat[\textbf{2D skeleton v.s. 3D skeleton.}  \label{tab-poseext-2Dvs3D}]{
		\tablestyle{8pt}{1.3}
		\begin{tabular}{ccc} \shline
			Input          	& GYM & NTU-60 \\ \shline
			Kinect-3D~\cite{zhang2012microsoft}   	& N.A.   & 89.4   \\ 
			DOPE-3D~\cite{weinzaepfel2020dope} 		& 76.3   & N.A.   \\ 
			VIBE-3D~\cite{kocabas2020vibe}			& 87.0   & N.A.   \\ \shline
			HRNet-2D~\cite{sun2019deep} 		& \textbf{92.0}   & \textbf{91.9}   \\ 
			MobileNet-2D~\cite{howard2017mobilenets} 	& \textbf{89.0}   & \textbf{90.2}   \\ \shline 
		\end{tabular}}
	\hspace{5mm}
	\subfloat[\textbf{Lifted 3D-pose doesn't help in recognition.  }  \label{tab-poseext-3Dlifting}]{
		\tablestyle{22pt}{1.3}
		\begin{tabular}{cc} 
			\shline
			Input       & GYM 	\\ 
			\shline
			DOPE~\cite{weinzaepfel2020dope}     	& 76.3  \\ 
			VIBE~\cite{kocabas2020vibe}			& 87.0  \\ 
			\shline
			FrameLift~\cite{martinez2017simple} 	& 90.0  \\ 
			VideoLift~\cite{pavllo20193d} 			& 90.2  \\ 
			\shline
			HRNet-2D~\cite{sun2019deep}			& 92.0	\\ 
			\shline
		\end{tabular}}}

	\vspace{2mm}
	\resizebox{\linewidth}{!}{
	\subfloat[\textbf{Pose Estimation: Top-Down v.s. Bottom-Up. } \label{tab-poseext-TopvsBottom}]{
		\tablestyle{6pt}{1.3}
		\begin{tabular}{ccc}
			\hline
			Pose Estimator      & COCO AP & NTU-60 \\ \shline
			HRNet (Top-Down)     & \textbf{0.746}   & \textbf{93.6}   \\ \hline
			HRNet (Bottom-Up)    & 0.654   & 93.0   \\ \hline
			Mobile (Top-Down) & 0.646   & 92.0   \\ \hline
			\multicolumn{3}{c}{} \\
		\end{tabular}} 
	\hspace{3mm}
	\subfloat[\textbf{Pose extracted with different boxes.} 
		\label{tab-poseext-GTBox}]{
		\tablestyle{6pt}{1.3}
			\begin{tabular}{cc} 
				\hline
				Human Proposals & GYM Mean-Top1 \\ \shline
				Detection & 75.8      \\ \hline
				Tracking  & 85.3      \\ \hline
				GT        & \textbf{92.0}      \\ \hline
				\multicolumn{2}{c}{} \\
			\end{tabular}}
	\hspace{3mm}
	\subfloat[\textbf{Coordinate v.s. Heatmap.}  
	\label{tab-poseext-coordvsheat}]{
		\tablestyle{6pt}{1.3}
			\begin{tabular}{cc}
				\shline
				Input Format & GYM Mean-Top1 \\ 
				\shline
				Coordinate-MobileNet & 90.7      \\
				Coordinate-HRNet & 93.2      \\ 
				\shline
				Heatmap-MobileNet    & 92.7      \\
				Heatmap-HRNet    & 93.6      \\ 
				\shline
		\end{tabular}}}
	\vspace{-4mm}
\end{table*}

\subsection{Ablation Study on Pose Extraction}
\label{sec-abl-pose-extraction}
This section discusses different alternatives that can be adopted in pose extraction to validate our choice. 
The input size for all 3D-CNN experiments is $ T\times H\times W = 48\times 56\times 56$.

\textbf{2D v.s. 3D Skeletons.}
We first compare the recognition performance of using 2D and 3D skeletons for action recognition. 
The 3D skeletons are either collected by sensors (NTU-60) or estimated with state-of-the-art 3D pose estimators based on RGB inputs~\cite{weinzaepfel2020dope,kocabas2020vibe} (FineGYM). 
For a fair comparison, we use MS-G3D~\cite{liu2020disentangling} (the current state-of-the-art GCN for skeleton-based action recognition) 
with the same configuration and training schedule for 2D and 3D keypoints and list the results in Table~\ref{tab-poseext-2Dvs3D}.
The estimated 2D keypoints (even low-quality ones) consistently outperform 3D keypoints (sensor collected or estimated) in action recognition.
Besides RGB-based 3D-pose estimators, we also consider the `lifting' approaches~\cite{martinez2017simple,pavllo20193d}, 
which directly `lift' 2D-pose (sequences) to 3D-pose (sequences).
We regress the 3D poses based on 2D poses extracted with HRNet, use the lifted 3D poses for action recognition.
The results in Table~\ref{tab-poseext-3Dlifting} indicate that such lifted 3D poses do not provide 
any additional information, performs even worse than the original 2D poses in action recognition.

\begin{table}[t]
	\captionsetup{font=small}
	\centering
	\caption{\textbf{Transferring Ability. } Skeleton representations learned on the large-scale Kinetics400
											 can transfer to downstream datasets well. Backbone parameters are frozen
											 for the `Linear' setting. }
	\vspace{-2mm}
	\label{tab-transfer}
	\resizebox{.7\columnwidth}{!}{
	\tablestyle{10pt}{1.3}
	\begin{tabular}{ccc}
		\shline
		Policy  & HMDB51 & UCF101 \\ 
		\shline
		Scratch  & 58.6   & 79.1  \\ 
		Linear   & 64.9   & 83.1  \\ 
		Finetune & \textbf{69.3}   & \textbf{87.0}  \\ 
		\shline
	\end{tabular}}
	\vspace{-5mm}
\end{table}

\textbf{Bottom-Up v.s. Top-Down.}
To compare the pose estimation quality of Bottom-Up and Top-Down approaches, 
we instantiate the two approaches with the same backbone (HRNet-w32). 
Besides, we also instantiate the Top-Down approach with the MobileNet-v2 backbone for comparison, 
which has a similar performance to HRNet (Bottom-Up) on COCO-validation.
We use extracted 2D poses to train a \emph{Pose-SlowOnly} on NTU-60. 
Table~\ref{tab-poseext-TopvsBottom} shows that
the performance of HRNet (Bottom-Up) on COCO-val is much worse than HRNet (Top-Down) and close to MobileNet (Top-Down). 
However, the Top-1 accuracy of HRNet (Bottom-Up) is much higher than MobileNet (Top-Down) and close to HRNet (Top-Down).
Although the potential of Bottom-Up should not be neglected, 
considering the better performance and faster inference speed (Top-Down runs faster when there aren't many persons in a frame),
we use Top-Down for pose extraction in this work.

\textbf{Interested Person v.s. All Persons. }
Many people may exist in a video, but not all of them are related to the interested action.
For example, in FineGYM, only the pose of the athlete is helpful, while other persons like the audience or referee are unrelated.
We compare using 3 kinds of person bounding boxes for pose extraction: \textbf{Detection}, \textbf{Tracking}(with Siamese-RPN ~\cite{li2018high}) and \textbf{GT} (with increasing prior about the athlete).  
In Table~\ref{tab-poseext-GTBox}, we see that the prior of the interested person is extremely important: even weak prior knowledge (1 GT box per video) can improve the performance by a large margin.

\textbf{Coordinates v.s. Heatmaps. }
Storing 3D heatmap volumes may take vast amounts of disk space. 
To be more efficient, we can save the 2D poses as coordinate-triplets (x, y, score)
and restore them to 3D heatmap volumes following the methods we introduced in Sec~\ref{sec-gen_volume}.
We conduct experiments on FineGYM to explore how much information is lost during the heatmap $\rightarrow$ coordinate compression.
In Table~\ref{tab-poseext-coordvsheat}, we see that for low-quality pose estimators, it leads to a 2\% drop in Mean-Top1.
For high-quality ones, the degradation is more moderate (only a 0.4\% Mean-Top1 drop). 
Thus we choose to store coordinates instead of 3D heatmap volumes.

\subsection{Multi-Modality Action Recognition Results on UCF101 and HMDB51}
\label{sec-mmar-ucfhmdb}

In Table~\ref{tab-volumevsaggr}, we train different PoseConv3D on UCF101 and HMDB51 from scratch.
In this section, we demonstrate that PoseConv3D can also take advantage of pretraining on large-scale datasets.
We adopt weights pretrained on Kinetics400	to initialize the PoseConv3D. 
Pretraining with skeleton data from the large-scale Kinetics400 benefits the downstream recognition tasks on smaller datasets, 
under both `Linear' and `Finetune' paradigms (Table~\ref{tab-transfer}).

We further compare PoseConv3D with previous state-of-the-arts of skeleton-based action recognition on UCF101 and HMDB51: 
PoTion~\cite{choutas2018potion} and PA3D~\cite{yan2019pa3d}. 
For a fair comparison, we fuse the skeleton-based predictions with I3D~\cite{carreira2017quo} predictions, 
instead of predictions from the more advanced OmniSource~\cite{duan2020omni}. 
Table~\ref{tab-ucfhmdbsota} shows that PoseConv3D not only 
outperforms other approaches by a large margin on skeleton-based action recognition,  
but also leads to better overall performance after fusing with predictions based on other modalities. 

\subsection{Using 3D Skeletons in PoseConv3D}

PoseConv3D takes stacked 2D skeleton keypoint heatmaps as input.
Assume only 3D skeletons are available, 
one can also use the 3D skeletons in PoseConv3D by projecting them to a 2D plane. 
The NTURGB+D dataset~\cite{Shahroudy_2016_NTURGBD} provides 
3D skeleton sequences collected by Microsoft Kinect v2 sensors~\cite{zhang2012microsoft}.
Besides, the dataset also includes the projection of 3D joints onto the 2D image coordinate systems.
We use the projected 2D skeletons of NTU-60 as the input for PoseConv3D and study the effect.

\begin{table}[t]
	\captionsetup{font=small}
	\centering
	\caption{\textbf{Comparison with state-of-the-art multi-modality action recognition approaches. }}
	\vspace{-2mm}
	\label{tab-ucfhmdbsota}
	\resizebox{.7\columnwidth}{!}{
	\tablestyle{8pt}{1.3}
	\begin{tabular}{ccc} \shline
	Method & HMDB51 & UCF101 \\ \shline
	I3D~\cite{carreira2017quo}   & 80.7   & 98.0   \\ \shline
	PoTion~\cite{choutas2018potion} & 43.7   & 65.2   \\
	PoTion + I3D     & 80.9   & 98.2   \\ \shline
	PA3D~\cite{yan2019pa3d}    & 55.3   & -      \\
	PA3D + I3D       & 82.1   & -      \\ \shline
	PoseConv3D       & \textbf{69.3}   & \textbf{87.0}   \\ 
	PoseConv3D + I3D & \textbf{82.7}   & \textbf{98.4}   \\ \shline
	\end{tabular}}
	\vspace{-2mm}
\end{table}

\begin{table}[t]
	\captionsetup{font=small}
	\centering
	\caption{\textbf{PoseConv3D with projected 2D poses.} We report the recognition performance of the joint model. }
	\label{tab-proj}
	\vspace{-2mm}
	\resizebox{.7\columnwidth}{!}{
	\tablestyle{7pt}{1.3}
	\begin{tabular}{cc}
		\shline
		Input + Method & Top-1 \\ 
		\shline
		2D-projection + MS-G3D~\cite{liu2020disentangling}  & 86.8    \\
		3D-skeleton + MS-G3D~\cite{liu2020disentangling}   & 88.8\footnotemark  \\	
		\shline
		2D-projection + PoseConv3D & \textbf{89.2}    \\ 
		\shline
	\end{tabular}}
	\vspace{-4mm}
\end{table}

Table~\ref{tab-proj} demonstrates the recognition performance of using projected 2D skeletons in PoseConv3D.
Using the projected 2D skeletons as input instead of the original 3D skeletons, 
there is a 2\% Top-1 accuracy drop for MS-G3D due to the information lost in 3D $\rightarrow$ 2D compression.
If both use 2D skeletons as input, PoseConv3D outperforms the GCN-based counterpart by 2.4\%, 
even surpasses the MS-G3D with 3D skeletons as input by 0.4\%, 
which indicates the great spatiotemporal modeling capability of 3D-CNN can compensate for the information lost in projection.

\footnotetext{We rerun the official code of MS-G3D to get this accuracy. }

\subsection{Practice for Group Activity Recognition}
\label{sec-groupractice}

\begin{figure}[t]
    \captionsetup{font=small}
    \includegraphics[width=\linewidth]{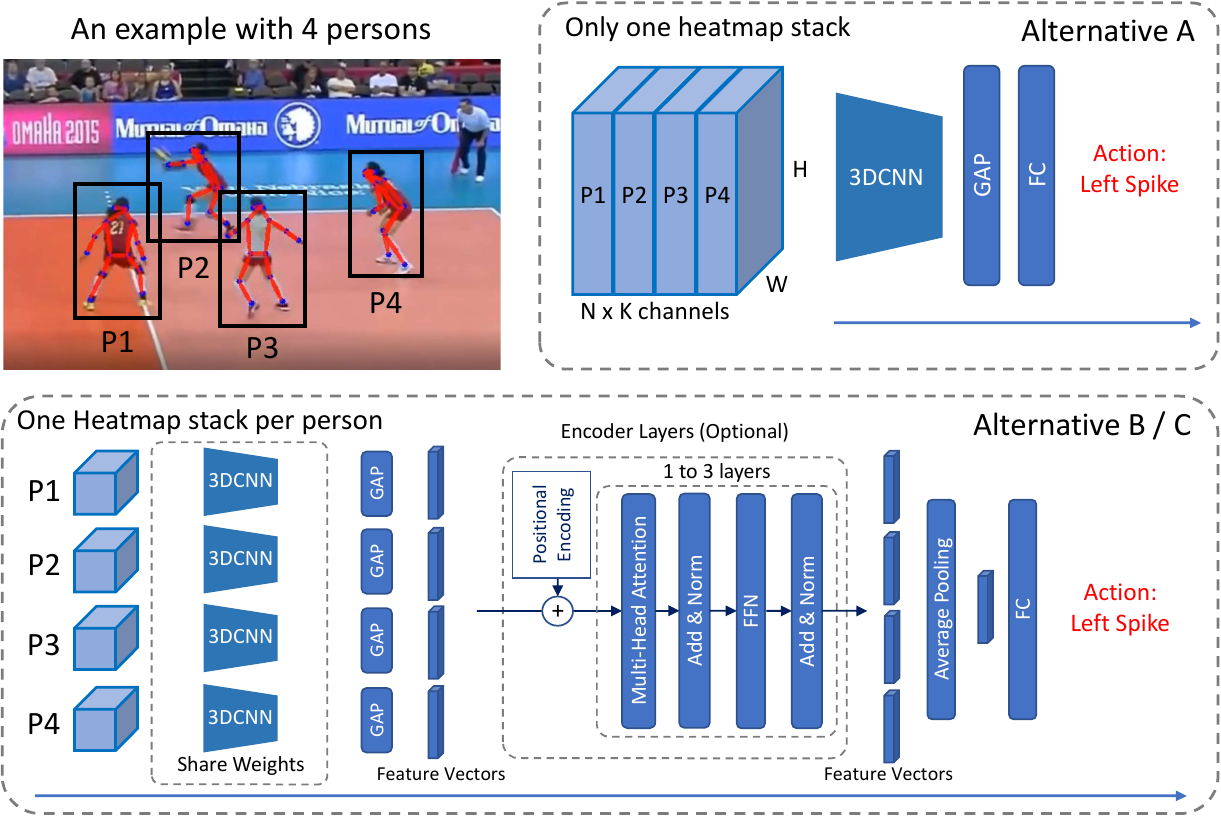}
	\vspace{-6mm}
    \caption{An illustration (with 4 persons) of the proposed three alternatives for group activity recognition. 
	Best viewed in 4x. }
	\vspace{-2mm}
    \label{fig}
\end{figure}

In experiments, we find that representing all people with a single heatmap volume is 
the best practice for group activity recognition with PoseConv3D.
On the Volleyball dataset, we have also explored three alternatives that process different persons' heatmaps separately:
\textbf{A.} For each joint, we allocate $N$ channels for $N$ persons. The PoseConv3D input then has $N\times K$ channels (instead of $K$); 
\textbf{B.} We generate a 3D heatmap volume ($K\times T\times H\times W$) for each person and use PoseConv3D (weights shared among $N$ persons) to extract the skeleton feature separately. 
We use average pooling to aggregate $N$ persons' features to a single feature vector;
\textbf{C.} On top of \textbf{B}, we insert several (1 to 3) encoder layers (from scratch or with \textbf{B} pre-training) before the average pooling for inter-person modeling.  
Figure~\ref{fig} provides an illustration of three alternatives.
For \textbf{A}, the high dimensional input leads to severe overfitting. The Top-1 accuracy is only 75.3\%. 
For \textbf{B, C}, despite the great amounts of computation ($> 13 \times$) consumed, 
the recognition performance is not satisfying. 
At best, \textbf{B}, \textbf{C} achieves 85.7\% and 87.9\% Top-1 on Volleyball, 
still much inferior to accumulating heatmaps (91.3\%). 
Accumulating heatmaps is a simple and relatively good solution for balancing complexity and effectiveness. 
More complex designs may lead to further improvements, which is left to future work. 

\subsection{Uniform Sampling for RGB-based recognition}
\label{sec-uniformrgb}

\begin{table}[t]\centering
	\captionsetup{font=small}
	\captionsetup[subffloat]{justification=centering,position=bottom}
	\caption{\textbf{Uniform sampling also works for RGB-based action recognition. }
		Alls results are for 10-clip testing, except the `uniform-16 (1c)', which uses 1-clip testing.}
	\vspace{-2mm}
	\label{tab-rgbuni}
	\resizebox{\columnwidth}{!}{
		\subfloat[\footnotesize \textbf{FineGYM.} ]{
			\tablestyle{6pt}{1.3}
			\begin{tabular}{cc}
				\shline
				Sampling   & Mean-Top1 \\ 
				\shline
				16x2       & 87.9     \\ 
				16x4       & 88.7     \\ 
				\shline
				uniform-16 (1c) & \textbf{91.1} \\
				uniform-16 & \textbf{91.6}     \\ 
				\shline
			\end{tabular}}
		\hspace{5mm}
		\subfloat[\footnotesize \textbf{NTU-60 (X-Sub)}]{
			\tablestyle{11pt}{1.3}
			\begin{tabular}{cc}
				\shline
				Sampling   & Top1 \\ 
				\shline
				16x2       & 94.9     \\
				16x4       & 95.1     \\ 
				\shline
				uniform-16 (1c) &  \textbf{95.7} \\
				uniform-16 & \textbf{96.1}     \\ 
				\shline
			\end{tabular}}} 
	\vspace{-5mm}
\end{table}

Based on the outstanding improvement by uniform sampling on skeleton action recognition,
we wonder if the strategy also works for RGB-based action recognition.
Thus we apply uniform sampling to RGB-based action recognition on NTU-60~\cite{Shahroudy_2016_NTURGBD} and GYM~\cite{shao2020finegym}.
We use SlowOnly-R50~\cite{feichtenhofer2019slowfast} as the backbone and set the input length as 16 frames.
From Table~\ref{tab-rgbuni}, 
we see that uniform sampling also outperforms fix-stride sampling by a large margin in RGB-based recognition on two datasets:
the accuracy of uniform sampling with 1-clip testing is better than the accuracy of fix-stride sampling with 10-clip testing.
We mainly attribute the advantage of uniform sampling to the highly variable video lengths in these two datasets.
On the contrary, we observe a slight accuracy drop on Kinetics400\footnote{
	In Kinetics400, most video clips are of the same temporal length: 10s.}
when applying uniform sampling: 
for SlowOnly-R50 with input length 8, the Top-1 accuracy drops from 75.6\% to 75.2\%.

\subsection{NTU-60 Error Analysis}

\begin{table}[t]
	\captionsetup{font=small}
	\centering
	\caption{\textbf{Top 5 confusion pairs of skeleton action recognition on NTU60 XSub. } 
	Multi-modality fusion with \emph{RGBPose}-Conv3D improves the recognition performance on confusion pairs by a lot.}
	\vspace{-2mm}
	\label{tab-confusing}
	\resizebox{\columnwidth}{!}{
	\tablestyle{8pt}{1.3}
	\begin{tabular}{cccc}
		\shline
		Action1       & Action2                   & $S_{\mathrm{Pose}}$ & $S_{\mathrm{RGB+Pose}}$ \\ 
		\shline
		Read          & Play with phone/tablet    & 67                     & 13                           \\ 
		Write         & Type on a keyboard        & 57                     & 20                           \\ 
		Write         & Play with phone/tablet    & 50                     & 5                            \\ 
		Take a selfie & Point to sth. with finger & 48                     & 10                           \\ 
		Read          & Write                     & 44                     & 24                           \\ 
		\shline
	\end{tabular}}
	\vspace{-3mm}
\end{table}

On NTU-60 X-Sub split, we achieve 94.1\% Top-1 accuracy with skeleton-based action recognition, 
which outperforms the current state-of-the-art result by 2.6\%. 
To further study the failure cases, 
we first define the confusion score $S$ of a pair of action classes $i$, $j$ as $S = n_{ij} + n_{ji}$
($n_{ij}$: number of videos of class $i$ but recognized as class $j$).
In NTU-60, there are 1770 pairs of action classes in total,
while we list the five most confusing pairs in Table~\ref{tab-confusing}.
Most failure cases are of these top-confusing pairs, \eg, 
over 27\% failure cases are of the top 5 confusion pairs. 
It is hard to distinguish these pairs of actions with human skeletons only.

Some confusing pairs can be resolved by exploiting other modalities such as RGB appearance. 
If the model successfully recognizes the keyboard, then it can distinguish typing from writing.
Table~\ref{tab-confusing} shows that, 
with multi-modality fusion in \emph{RGBPose}-Conv3D, 
the recognition performance on those confusing pairs improves a lot.

\subsection{Why skeleton-based pose estimation performs poorly on Kinetics400}
\label{sec-error}

\begin{figure}[t]
	\captionsetup{font=small}
	\centering
	\includegraphics[width=\columnwidth]{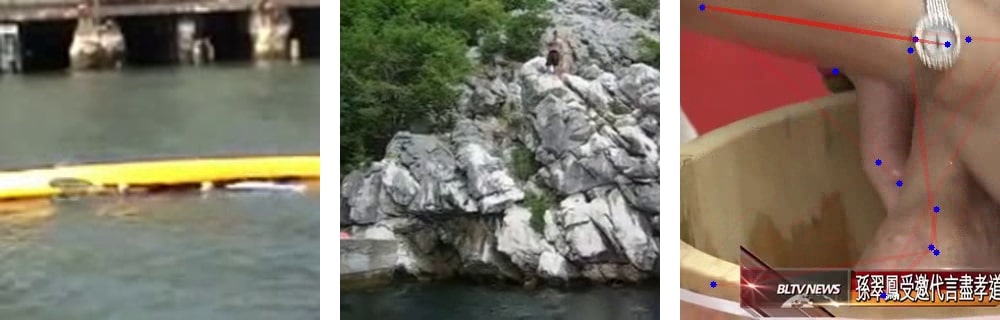}
	\vspace{-6mm}
	\captionof{figure}{\textbf{Problems in Kinetics400 Pose Extraction. } 
		Left: Human missing in action `kayaking'.
		Middle: Human skeleton is too small to be recognized in action `diving cliff'.
		Right: Only human parts appear, the pose estimator fails (`washing feet'). 
	}
	\vspace{-5mm}
	\label{fig-kineticsvis2}
\end{figure}

\begin{table}[t]
	\captionsetup{font=small}
	\centering
	\caption{\textbf{Mean class accuracy on the Kinetics-Motion subset.}}
	\vspace{-2mm}
	\label{tab-kineticsmotion}
	\resizebox{.65\columnwidth}{!}{
	\tablestyle{8pt}{1.25}
	\begin{tabular}{cc}
		\hline
		Method & Top1 Acc \\ \shline
		Swin-L~\cite{liu2021video} 	 & 92.7    \\ \shline
		ST-GCN~\cite{yan2018spatial}    & 72.0    \\ \hline
		PoseConv3D		         		 & 81.9    \\ \shline
		Swin-L + PoseConv3D		 		 & \textbf{94.7}    \\ \shline
	\end{tabular}}
	\vspace{-6mm}
\end{table}

PoseConv3D with high-quality 2D skeletons improves the Top-1 accuracy of skeleton-based action recognition 
on Kinetics400 from 38.0\% to 47.7\%.
However, the accuracy on Kinetics400 is still far below the accuracies on other datasets. 
Besides the difficulties mentioned in Sec~\ref{sec-vis},
two more problems will degrade the quality of extracted skeleton sequences (Figure~\ref{fig-kineticsvis2}):
1. Since Kinetics400 is not human-centric, human skeletons are missing or hard to recognize in many frames.
2. For the same reason, only small parts of humans appear in many frames, while the pose estimators are easy to fail in this scenario.

We also report the mean class accuracy on Kinetics-Motion~\cite{yan2018spatial} in Table~\ref{tab-kineticsmotion}, 
which contains 30 action classes in Kinetics that are strongly related to body motions.
The accuracy of skeleton-based action recognition is much higher on this subset, increasing from 47.7\% to 81.9\%. 
When combined with the state-of-the-art RGB predictions, the improvement is much more significant, increasing from 0.6\% to 2.0\%. 
However, the skeleton-based performance is still far behind the state-of-the-art RGB-based action recognition method~\cite{liu2021video}, 
which achieves 92.7\% mean class accuracy on Kinetics-Motion.
The inferior recognition performance indicates that there still needs more future work for skeleton-based action recognition in the wild.

\vspace{-3mm}
{\small
\bibliography{egbib}
\bibliographystyle{ieee_fullname}
}

\end{document}